\documentclass[twoside,11pt]{article}

\usepackage{jmlr2e_arxiv}
\graphicspath{{fig/}}
\usepackage{algorithm}
\usepackage{algorithmic}
\usepackage{amsmath}

\newcommand{\argmin}[1]{\underset{#1}{\operatorname{arg}\,\operatorname{min}}\;}
\usepackage{tablefootnote}
\usepackage{subcaption} 
\usepackage{color}

\ShortHeadings{A particle-based variational approach to BNMF}{Masood and Doshi-Velez}
\firstpageno{1}

\begin{document}

\title{A particle-based variational approach to Bayesian Non-negative Matrix Factorization}

\author{\name Muhammad A Masood \email masood@g.harvard.edu \\
       \addr Harvard John A. Paulson \\
       School of Engineering and Applied Science\\
       Cambridge, MA 02138, USA
       \AND
       \name Finale Doshi-Velez \email finale@seas.harvard.edu \\
        \addr Harvard John A. Paulson \\
       School of Engineering and Applied Science\\
       Cambridge, MA 02138, USA}

\maketitle

\begin{abstract}%   <- trailing '%' for backward compatibility of .sty file
Bayesian Non-negative Matrix Factorization (NMF) is a promising approach for understanding uncertainty and structure in matrix data. However, a large volume of applied work optimizes traditional non-Bayesian NMF objectives that fail to provide a principled understanding of the non-identifiability inherent in NMF-- an issue ideally addressed by a Bayesian approach. Despite their suitability, current Bayesian NMF approaches have failed to gain popularity in an applied setting; they sacrifice flexibility in modeling for tractable computation, tend to get stuck in local modes, and require many thousands of samples for meaningful uncertainty estimates. We address these issues through a particle-based variational approach to Bayesian NMF that only requires the joint likelihood to be differentiable for tractability, uses a novel initialization technique to identify multiple modes in the posterior, and allows domain experts to inspect a `small' set of factorizations that faithfully represent the posterior. We introduce and employ a class of likelihood and prior distributions for NMF that formulate a Bayesian model using popular non-Bayesian NMF objectives. On several real datasets, we obtain better particle approximations to the Bayesian NMF posterior in less time than baselines and demonstrate the significant role that multimodality plays in NMF-related tasks. 

\end{abstract}

\section{Introduction}

The goal of non-negative matrix factorization (NMF) is to find a rank-$R$ factorization for a non-negative data matrix $X$ ($D$ dimensions by $N$ observations) into two non-negative factor matrices $A$ and $W$. Typically, the rank $R$ is much smaller than the dimensions and observations ($R \ll D,N$).

\begin{equation*}
X \approx AW \quad | \quad \quad  X \in \mathbb{R}_{+}^{D \times N}, \quad A \in \mathbb{R}_{+}^{D \times R},  \quad W \in \mathbb{R}_{+}^{R \times N}
\end{equation*}

The linear additive structure of these non-negative factor matrices makes NMF a popular unsupervised learning framework for discovering and interpreting latent structure in data. Each observation in the data $X$ is approximated by an additive combination of the $R$ columns of $A$ with the combination weights given by the column of $W$ corresponding to that observation. In this way, the basis matrix $A$ provides a part-based representation of the data and the weights matrix $W$ provides an $R$-dimensional latent representation of the data under this part-based representation. 

Applications of NMF include understanding protein-protein interactions \citep{Greene}, topic
modeling \citep{roberts2016navigating}, hyperspectral unmixing \citep{bioucas2012hyperspectral}, polyphonic music transcription \citep{smaragdis2003non}, discovering molecular
pathways from genomic samples \citep{brunet2004metagenes}, and summarizing activations of a neural network for greater interpretretability \citep{olah2018the}. 

NMF is known to be a difficult problem: in the absence of noise, NMF is NP-hard \citep{vavasis2009complexity} and NMF objectives in general are non-convex. In addition, analysis and interpretation of latent structure in a dataset via NMF is affected by the the possibility that several non-trivially different pairs of $A, W$ may reconstruct the data $X$ equally well. This non-identifiability of the NMF solution space has been studied in detail in the theoretical literature \citep{pan2016characterization,donoho2003does,arora2012computing,intersectingfaces15,bhattacharyya2016non}, and is of practical concern as well. \citet{Greene} use ensembles of NMF solutions to model chemical interactions, while \citet{roberts2016navigating} conduct a detailed empirical study of multiple optima in the context of extracting topics from large corpora. 

Bayesian approaches to NMF promise to characterize this parameter uncertainty in a principled manner by solving for the posterior
$p(A,W|X)$ given priors $p(A)$ and $p(W)$ and likelihood $p(X | A, W)$ e.g. \citet{schmidt2009bayesian,
  moussaoui2006separation}.  Having a representation of uncertainty in the parameters of the factorizations can assist with the proper interpretation of the factors, allowing us to place low or high confidence on parameters of the
factorization. However, computational tractability of inference in Bayesian models is a concern that limits the application of the Bayesian approach. Additionally, the uncertainty estimates obtained from current Bayesian methods are often of limited use: variational approaches
(e.g. \citet{paisley2015bayesian, hoffman2015structured}) typically
underestimate uncertainty and fit to a single mode; sampling-based
approaches (e.g. \citet{schmidt2009bayesian, moussaoui2006separation})
also rarely switch between multiple modes and require many thousands of samples for meaningful uncertainty estimates.  

As a result of the limitations of current Bayesian approaches, practitioners tend to rely on non-Bayesian approaches to understand structure in matrix data and explore uncertainty in NMF parameters. For example, \citet{Greene, roberts2016navigating,brunet2004metagenes} use random restarts to find multiple solutions. Random restarts have no Bayesian interpretation (depending on the basins of attraction of each mode), but they do often find multiple optima in the objective that can be used to understand and interpret the data. Despite its usefulness, the random restarts approach lacks the principled nature of a Bayesian treatment.

We introduce a Bayesian NMF framework that incorporates the benefits of non-Bayesian approaches into the principled Bayesian context by constructing a discrete distribution over the space of factorizations where the probability masses are optimized under the Stein variational objective  \citep{ranganath2016operator}. In addition, we provide a novel initialization strategy for NMF that uses transfer learning to provide high-quality and diverse initializations for NMF that reduce optimization time for NMF algorithms and find multiple optima in the NMF objective. Finally, this work also introduces a class of likelihood and prior models for Bayesian NMF that easily allow practitioners to re-formulate their problems (using traditional non-Bayesian NMF objectives) as Bayesian. Once the Bayesian model is chosen, a set of factorizations from high-likelihood regions of the posterior are found by combining our novel initialization technique with state-of-the-art non-Bayesian NMF algorithms. Subsequently, probability masses are assigned to each factorization such that the divergence from the posterior as measured by the Stein variational objective is minimized.
 
 We demonstrate through multiple real datasets that: 

\begin{itemize}
\item Our particle-based posterior approximation framework consistently outperforms baselines in terms of approximation quality and runtime. 
\item Our framework produces diverse sets of particles from multiple modes of the posterior landscape that have distinct interpretations, exhibit variability in performance on downstream tasks, and improve user ability to inspect and understand the full solution space. 
\end{itemize}

\section{Inference Setting}

We seek to approximate the Bayesian NMF posterior $p(\theta | X)$ with a discrete variational distribution $q(\theta | \theta_{1:M}, w_{1:M})$ that has $M$ different point-masses $\theta_{m}$. Each $\theta_m$ represents a different NMF solution's full set of parameters: $\theta_m = \{A_m, W_m\}$, and is assigned probability mass $w_m$. The functional form of the variational distribution is given by:

\begin{equation}
\label{eqn:discrete_q}
\begin{split}
p(\theta | X) & \approx q(\theta | \theta_{1:M}, w_{1:M}) = \sum_{m = 1}^M w_m \delta(\theta - \theta_m) \quad  \\
& s.t \quad w_{1:M} \in \Delta^{M-1}, \quad \text{where} \quad \theta_m = \text{vec}[A_m^T, W_m]
\end{split}
\end{equation}
where $\delta$ is the Dirac delta distribution and $\Delta^{M-1}$ is the probability simplex in $\mathbb{R}^M$. The task of effectively constructing this discrete approximation depends on producing high-quality, diverse factorization collections $\theta_{1:M}$ and determining their associated weights $w_{1:M}$.

The approximation quality between the discrete distribution $ q(\theta | \theta_{1:M}, w_{1:M})$ and the posterior distribution $p(\theta | X)$ of interest can be quantified using recent advances in Stein discrepancy evaluation with kernels
\citep{liu2016two,chwialkowski2016kernel,gretton2006kernel,liu2016kernelized,gorham2015measuring}. While the popular Kulback-Leibler divergence requires comparing the ratio of probability densities or probability masses, the Stein discrepancy can be used to compare a particle-based collection defined by probability masses with a continuous target distribution. Equation~\ref{eqn:opt_obj} summarizes the objective of our particle-based variational approach: we minimize the Stein discrepancy $\mathbb{S}_p(q)$ with respect to the Bayesian NMF posterior $p(\theta | X)$ of a discrete variational distribution $q(\theta | \theta_{1:M},w_{1:M})$. In Section~\ref{sec:bnmf_approximation}, we introduce our main algorithm to solve this objective efficiently (Algorithm~\ref{alg:posterior_Q}). 

\begin{equation}
\label{eqn:opt_obj}
q^*(\theta | \theta_{1:M}, w_{1:M} ) = \text{argmin}_{ \theta_{1:M}, w_{1:M} } \mathbb{S}_{p(\theta | X)}(q(\theta | \theta_{1:M}, w_{1:M} ) ) \quad s.t. \quad w_{1:M} \in \Delta^{M-1}
\end{equation}

\section{Background}
\paragraph{Bayesian Non-negative Matrix Factorization}

In Bayesian NMF, we define priors $p(A)$ and $p(W)$ and a likelihood
$p(X|A,W)$ and seek to characterize the posterior $p(A,W|X)$. These are related by Bayes' rule:

\begin{equation*}
p(A,W|X) = \frac{p(X|A,W)p(A)p(W)}{p(X)}
\end{equation*}
There exist many options for the choice of prior and likelihood (e.g.,
exponential-Gaussian, \citep{paisley2015bayesian,schmidt2009bayesian},  Gamma Markov chain priors \citep{dikmen2009unsupervised} and volume-based priors \citep{arngren2011unmixing}). The likelihood and prior choices are often sought in order to have good computational properties (e.g. the resulting partial conjugacy of exponential-Gaussian model). One feature of our work is that we do not require convenient priors for the inference task. 

\paragraph{Transfer Learning}
The field of transfer learning aims to leverage models and inference
applied to one problem to assist in solving related problems.  It is
of practical value because there may be an abundance of data and
computational resources for one problem but not another (see
\citet{pan2010survey} for a survey).  In this work, we shall use the
solutions to Bayesian NMF from small, synthetic problems to quickly
solve much larger NMF problems.

\paragraph{Stein discrepancy}
\label{sec:stein}

The Stein discrepancy $\mathbb{S}_p(q)$ is a divergence
from distributions $q(\theta)$ to $p(\theta)$ that only requires sampling from the variational distribution $q(\theta)$ and evaluating the score function of the target distribution $p(\theta)$. The Stein discrepancy is computed over some class of test functions $f \in \mathcal{F}$ and satisfies the closeness property for operator variational inference  \citep{ranganath2016operator}: it is non-negative in general and zero only for some equivalence class of distributions $q \in \mathcal{Q}_0$. For a rich enough function class, the only distribution for which the Stein discrepancy is zero is the distribution $p$ itself. The approximation quality between the discrete distribution $ q(\theta | \theta_{1:M}, w_{1:M})$ and the posterior distribution $p(\theta | X)$ of interest can be analytically computed using recent advances in Stein discrepancy evaluation with kernels \citep{liu2016two,chwialkowski2016kernel,gretton2006kernel,liu2016kernelized,gorham2015measuring}. 
The Stein discrepancy is related to the maximum mean
discrepancy (MMD): a discrepancy that measures the worst-case deviation between expectations of functions $h \in \mathcal{H}$ under $p$ and $q$ \citep{gretton2006kernel}. 

\begin{equation*}
\text{MMD}(\mathcal{H},q,p) := \sup_{h \in \mathcal{H}}
\mathbb{E}_{\theta \sim q} h (\theta) - \mathbb{E}_{\theta' \sim p} h (\theta')
\end{equation*}

By applying the Stein operator $\mathcal{T}_p$ corresponding to the distribution $p$, the function space $\mathcal{H}$ is transformed into another function space $\mathcal{T}_p(\mathcal{H}) = \mathcal{F}$. Expectations under $p$ of any $f \in \mathcal{F}$ are zero, i.e. $\mathbb{E}_{\theta' \sim p} f (\theta')) = 0$. This property of the Stein operator is of particular interest when the distribution $p$ is intractable because evaluating the Stein discrepancy does not require expectations over $p$ and the Stein operator $\mathcal{T}_p$ only depends on the unnormalized distribution via the score function $\nabla_\theta \log p(\theta)$. 

\begin{equation*}
\mathbb{S}_p(\mathcal{H}, q)  :=   \sup_{f \in \mathcal{T}_p(\mathcal{H})}
(\mathbb{E}_{\theta \sim q} f(\theta))^2 
\end{equation*}

In this work, we use a kernalized form of the Stein discrepancy. For every positive definite kernel $k(\theta, \theta')$, a unique Reproducing Kernel Hilbert Space (RKHS) $\mathcal{H}$ is defined. \citet{chwialkowski2016kernel} showed that the Stein operator applied to an RKHS defines a modified positive definite kernel $\mathcal{K}_p$ given by:
\begin{equation}
\label{eqn:stein_kernel}
\begin{split}
\mathcal{K}_p(\theta,\theta') & = \nabla_{\theta} \log p(\theta)^T \nabla_{\theta'} \log p(\theta') k(\theta,\theta') \\
& + \nabla_{\theta'} \log p(\theta')^T \nabla_\theta k(\theta,\theta')  \\
& + \nabla_{\theta} \log p(\theta)^T \nabla_{\theta'} k(\theta,\theta') \\
& +  \sum_{i = 1}^d \frac{\partial ^2 k(\theta,\theta')}{\partial \theta_i \partial \theta'_i}
\end{split}
\end{equation}
Finally, the Stein discrepancy is simply the expectation
of the modified kernel $\mathcal{K}_p$ under the joint distribution of two independent variables $\theta$, $\theta' \sim q$.

\begin{equation*}
\mathbb{S}_p(q) = \mathbb{E}_{\theta,\theta' \sim q} \mathcal{K}_p(\theta,\theta')
\label{eqn:stein_definition}
\end{equation*}
For a discrete distribution over $\theta_{1:M}$ with probability masses $w_{1:M}$ (of the form in equation~\ref{eqn:discrete_q}), this can be evaluated exactly \citep{liu2016black} as:
 \begin{equation}
 \begin{split}
\mathbb{S}_p(q) & =  \sum_{i,j = 1}^M w_iw_j \mathcal{K}_p(\theta_i,\theta_j) \\
& = \mathbf{w}^T \mathbf{K} \mathbf{w} \\
\end{split}
\label{eqn:s_obj}
 \end{equation}

The (pure) quadratic form $\mathbf{w}^T \mathbf{K} \mathbf{w}$ is a reformulation where $\mathbf{K} \in \mathbb{R}^{M \times M}$ is the (positive definite) pairwise kernel matrix with entries $\mathbf{K}_{ij} =  \mathcal{K}_p(\theta_i,\theta_j)$ and the probability masses $w_{1:M}$ are embedded into a vector $\mathbf{w} \in \mathbb{R}^{M \times 1}$. Our particle-based variational objective (equation~\ref{eqn:opt_obj}) simplifies to the form in equation~\ref{eqn:s_obj}. In Section ~\ref{sec:bnmf_approximation}, we will provide a method for estimating $\theta_{1:M}$ and $w_{1:M}$ for the Bayesian NMF problem.

\section{Approach}
\label{sec:bnmf_approximation}

In this section, we present a procedure to approximate the Bayesian NMF posterior with a discrete distribution over factorizations $\theta_{1:M}$ (summarized in Algorithm~\ref{alg:posterior_Q}). NMFs $\theta_{1:M}$ are found by running state-of-the-art (non-Bayesian) algorithms that are initialized using our novel transfer-based technique (Section~\ref{sec:transfer}). Given $\theta_{1:M}$, we apply out of the box convex optimization tools to find the optimal weights $w_{1:M}$ for representing the posterior i.e. minimize Stein discrepancy. 

\subsection{Learning factorization parameters $\theta_{1:M}$ via Transfer Learning}
\label{sec:transfer}

A natural approach to finding the factorization parameters $\theta_{1:M}$ is to optimize for them directly via the variational objective (equation~\ref{eqn:s_obj}), but we demonstrate this approach suffers from getting stuck in poor local optima and prohibitive computational requirements (see Section~\ref{sec:results}).  Since the quality of the variational approximation is determined solely by the value of the variational objective under a given set of parameters $\theta_{1:M}$, $w_{1:M}$, we are free to employ any technique that produces a suitable collection $\theta_{1:M}$. 

To explore multi-modal spaces (such as the Bayesian NMF posterior), random restart-based approaches\footnote{Random Restarts involve repeating an optimization procedure with different starting points that are independently sampled.} are known to introduce diversification to overcome the problem of converging to a single local mode \citep{gendreau2010handbook}. Specialized (non-Bayesian) optimization algorithms for NMF \citep{lee2001algorithms, lin2007projected, hsieh2011fast} are widely used in applied settings to produce single factorization parameters $\theta_m$. Since the convergence of these algorithms is dependent on initializations, random restarts can be used to discover multiple modes. That said, random restarts compound
computational concerns related to the slow convergence rates of many
(non-Bayesian) NMF algorithms---hence also a literature for
initializations that can speed up convergence
(\citet{salakhutdinov2002convergence}, \citet{wild2004improving},
\citet{xue2008clustering}, \citet{boutsidis2008svd}). 

\begin{figure}
\begin{centering}
\includegraphics[width=0.45\textwidth]{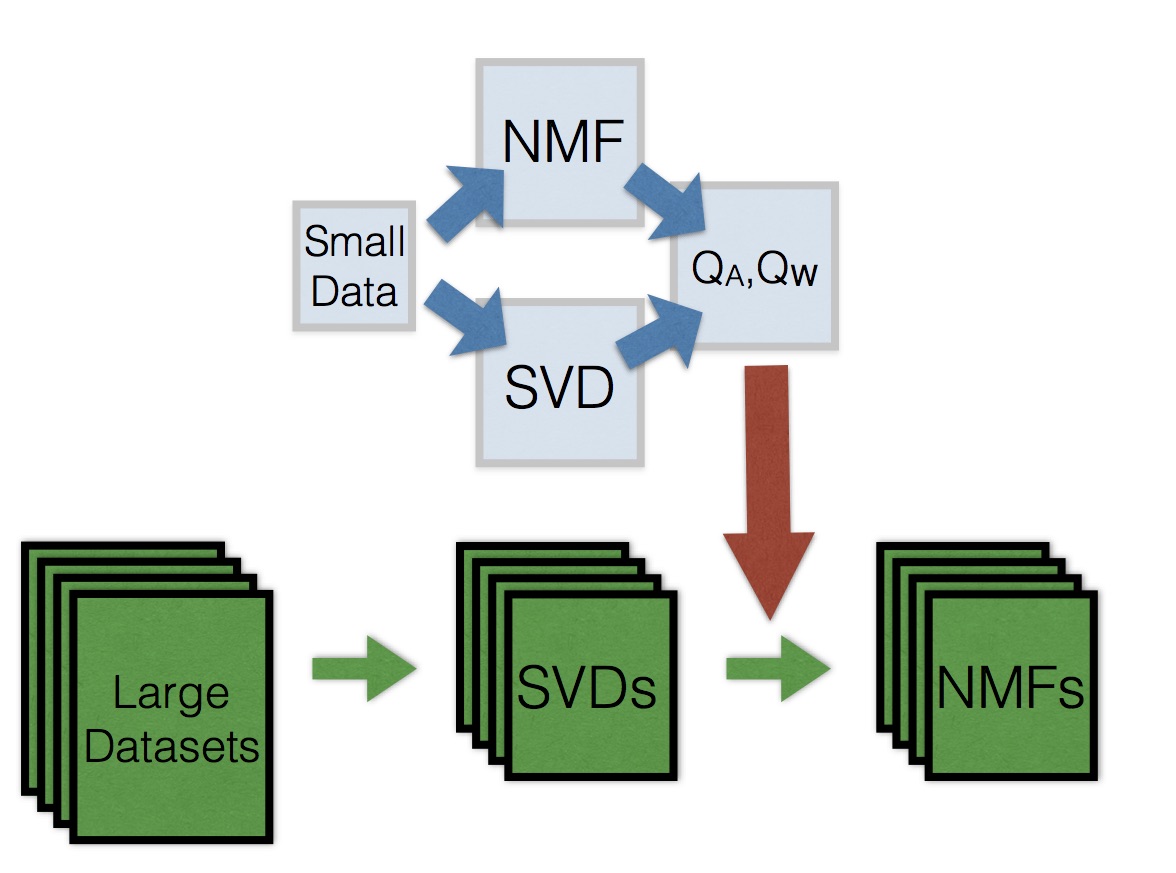}
\caption{A schematic of the transfer learning procedure for NMF: A small dataset is used to learn transformation matrices $Q_A,Q_W$. We then apply these transformation matrices to multiple larger datasets (with any number of dimensions or observations) using its SVD to obtain a transfer-based initialization.}
\label{fig:QT}
\end{centering}
\end{figure}

In this section, we introduce an initialization technique (which we will call $Q$-Transform) to speed-up, as compared to random restarts, the process of finding a diverse set of factorizations from high-density regions of the posterior. Initializations are determined by transforming the low-rank subspace of the singular value decomposition (SVD) using transformation matrices discovered via synthetic datasets. Figure~\ref{fig:QT} shows a schematic illustrating the idea that transformation matrices $Q_A,Q_W$ generated from one dataset can be applied to another dataset. In the context of this work, we repeatedly apply NMF algorithms to obtain a collection of diverse NMFs. The overall speedup due to $Q$-Transform is equal to the time advantage due to a single $Q$-Transform initialization, multiplied by the number of factorizations we compute ($M$). To explain our $Q$-Transform procedure, we first define the subspace transformation matrices, then describe the method for generating transformation matrices $Q_A, Q_W$ using synthetic data, and applying them to real data sets (transfer learning).

\begin{algorithm}[t]
   \caption{Particle-based Variational Inference for Bayesian NMF using $Q$-Transform}
   \label{alg:posterior_Q}
\begin{algorithmic}
   \STATE {\bfseries Input:} Data $\{X\}$, Rank $\{R_{\text{NMF}}\}$, \# Factorizations $M$
   \STATE {\bfseries Step 1:} Perform $M$ repetitions of Algorithm \ref{alg:Gen-Q} to get matrices $\{Q_A^m, Q_W^m\}_{m = 1}^M$ or re-use them if previously constructed
   \STATE {\bfseries Step 2:} Apply $Q$-Transform (Algorithm \ref{alg:Apply-Q}) to get Initializations $\{A_0^m, W_0^m\}_{m= 1}^M$
    \STATE {\bfseries Step 3:} Apply NMF algorithm to get Factorizations $\{A^m, W^m\}_{m = 1}^M$
    \STATE{\bfseries Step 4:}  Apply Algorithm~\ref{alg:eval_framework} using a given Bayesian NMF model to get weights $\{w^m\}_{m=1}^M$ for approximate posterior
        \STATE {\bfseries Output:} Discrete NMF Posterior $\{w^m, A^m,W^m\}_{m = 1}^M$ 
 \end{algorithmic}
\end{algorithm}

\paragraph{Subspace Transformations $Q_A,Q_W$ relating SVD and NMF.}
A low dimensional approximation for the data $X$ can be obtained via the top $R_\text{SVD}$ vectors of the SVD $A_{\text{SVD}}, W_{\text{SVD}}$. An NMF $A,W$ of rank $R_{T}$ (which may be different from $R_{\text{SVD}}$) also leads to an approximation of the data. The NMF factors are interpretable due to the non-negativity constraint whereas the SVD factors typically violate non-negativity. However, both approaches describe low dimensional subspaces that can be used to understand and approximate the data. These subspaces are the same when $R_{\text{SVD}} = R_{T}$ and the NMF is exact (i.e. $X = AW$; corresponds to Type I non-identifiability in
\citet{pan2016characterization}). Under these conditions, there exist
transformation matrices $Q_A , Q_W$ to obtain the non-negative
basis and weights in terms of the singular value decomposition matrices:

\begin{equation*}
\begin{split}
\text{If} \quad X = A_{\text{SVD}}W_{\text{SVD}} =  AW \quad \text{then} \\
 A = A_{\text{SVD}} Q_A \quad  W = Q_WW_{\text{SVD}}
\end{split}
 \end{equation*} 

When these conditions are not met, we may still expect the subspaces to be similar and expect that there exist transformation matrices $Q_A \in \mathbb{R}^{R_{\text{SVD}}\times R_{T}}, Q_W  \in \mathbb{R}^{R_{T}\times R_{\text{SVD}}}$ to yield approximations of the NMF factorizations. 

\begin{equation*}
 A_Q = A_{\text{SVD}}Q_A  \approx A \quad  W_Q = Q_WW_{\text{SVD}} \approx W 
 \end{equation*} 

The matrices $A_Q \in \mathbb{R}^{D \times R_{T}} , W_Q \in \mathbb{R}^{R_{T} \times N}$ approximate the NMF matrices but adjustments may be required to ensure the dimensions are equal to those of the factorization sought, and that the entries are non-negative before they can be used as an initialization $A_0, W_0$ (see Algorithm~\ref{alg:Init_Adjust}). 

\paragraph{Application of Transformations $Q_A, Q_W$ for NMF Initialization:}
If we have already computed an NMF $A,W$ for a dataset $X$, appropriate transforms $Q_A, Q_W$ can be computed by relating the SVD factors $A_{\text{SVD}}, W_{\text{SVD}}$ to $A,W$ (e.g. via linear least squares). This approach gives useful transformation matrices but provides no advantage since the NMF has already been computed. However, if we have a set of $Q_A, Q_W$ from one dataset, can we use them for another?

In the following, we shall generate these transforms using random restarts on small, synthetic datasets $X_s$ that follow some generative model for NMF, where we can run NMF algorithms quickly and solve for $Q_A,Q_W$ (Algorithm~\ref{alg:Gen-Q}). Multiple pairs of transformation matrices can be obtained by repeating Algorithm~\ref{alg:Gen-Q} with different random initializations to compute NMF of the synthetic data $X_s$, or by generating new synthetic datasets. We find that knowledge from these transformations $Q_A,Q_W$ can be transferred to real datasets by re-using them to relate the top SVD factors of other datasets to high quality, approximately non-negative factorizations. Since the transformations $Q_A, Q_W$ act on the inner dimensions (columns of $A_{\text{SVD}}$ and rows of $W_{\text{SVD}}$), we emphasize that they can be applied to new datasets with any number of dimensions $D$ and number of observations $N$. Given the top SVD factors of a new dataset $A_\text{SVD}, W_\text{SVD}$, we apply the $Q$-Transform (Algorithm~\ref{alg:Apply-Q}) which multiplies SVD factors by the $Q_A, Q_W$ matrices and adjusts entries of  $A_{\text{SVD}}Q_A$ and  $Q_WW_{\text{SVD}}$ using Algorithm~\ref{alg:Init_Adjust} to obtain initializations $A_0,W_0$ that can be used for any standard NMF algorithm (e.g. \citet{cichocki2009fast}, \citet{fevotte2011algorithms}) to give us factorization parameters $\theta_m = \text{vec}[A_m^T, W_m]$.  

We provide a publicly available demonstration of Algorithms~\ref{alg:Gen-Q} and ~\ref{alg:Apply-Q} in python to show the effectiveness of the $Q$-Transform initializations\footnote{https://github.com/dtak/Q-Transfer-Demo-public-/}. 

\begin{algorithm}[t]
   \caption{Generate $Q$-Transform Matrices}
   \label{alg:Gen-Q}
\begin{algorithmic}
   \STATE {\bfseries Input:} Synthetic Data $\{X_s\}$, SVD Dimension $\{R_{\text{SVD}}\}$, Transfer Dimension $\{R_{T}\}$
   \STATE $A_{\text{SVD}}, W_{\text{SVD}}$ $\leftarrow$ Compute top $R_{\text{SVD}}$ SVD of $X_s$
   \STATE $A_{\text{NMF}}, W_{\text{NMF}}$ $\leftarrow$ Compute rank-$R_{T}$ NMF of $X_s$ using random initialization
   \STATE $Q_A = \argmin{Q} \| A_{\text{NMF}} - A_{\text{SVD}}Q \|_{F} $  via linear least squares% $\quad$ $Q_W \in \mathbb{R}^{K_{\text{SVD}} \times K_{T}}$
     \STATE $Q_W = \argmin{Q} \| W_{\text{NMF}} - QW_{\text{SVD}} \|_{F} $  via linear least squares %$\quad$ $Q_A \in \mathbb{R}^{K_{T} \times K_{\text{SVD}}}$
   \STATE {\bfseries Output:} $Q_A$,  $Q_W$ 
\end{algorithmic}
\end{algorithm}

\begin{algorithm}[t]
   \caption{Apply $Q$-Transform}
   \label{alg:Apply-Q}
\begin{algorithmic}
   \STATE {\bfseries Input:} Real Data $\{X\}$, SVD Rank $\{R_{\text{SVD}}\}$, NMF Rank $\{R_{\text{NMF}}\}$, Transformation Matrices $\{Q_A, Q_W\}$
   \STATE $A_{\text{SVD}}, W_{\text{SVD}}$ $\leftarrow$ Compute top $R_{\text{SVD}}$ SVD of $X$
   \STATE $A_0, W_0 \leftarrow \text{Initialization Adjustment}(A_{\text{SVD}}Q_A,Q_WW_{\text{SVD}}, R_{\text{NMF}} )$ 
   \STATE {\bfseries Output:} $A_0$,  $W_0$ 
\end{algorithmic}
\end{algorithm}

\begin{algorithm}
\caption{Initialization Adjustment}
\label{alg:Init_Adjust}
\begin{algorithmic}
   \STATE {\bfseries Input:} Approximation matrices $\{A_Q, W_Q\}$, NMF Rank $\{R_{\text{NMF}}\}$
   \STATE {$\tilde{A_0} \leftarrow \text{Absolute Value} (A_Q)$}
   \STATE {$\tilde{W_0} \leftarrow \text{Absolute Value} (W_Q)$}
   \STATE Transfer Rank $R_{T}$ = $\#$ Columns of $A_Q$
   \IF  {NMF Rank $R_{\text{NMF}} >$ Transfer Rank $R_{T}$}
   \STATE $r = R_{\text{NMF}} - R_{T}$
   \STATE Pad $\tilde{A_0}, \tilde{W_0}$ with matrices $M_{D\times r}$ and $M_{N \times r}$ having small random entries so that initializations are the correct dimensions and matrices $M_{D\times r}, M_{N \times r}$ have little effect of the product $\tilde{A_0}\tilde{W_0}$.
   \STATE {$A_0 \leftarrow [\tilde{A_0}, M_{D\times k}]$}
   \STATE {$W_0 \leftarrow [\tilde{W_0}^T,M_{N \times k}]^T$}
   \ELSIF {NMF Rank $R_{\text{NMF}} < $ Transfer Rank $R_{T}$}
   \STATE Pick the top $R_{\text{NMF}}$ columns of $\tilde{A_0}$ and rows of $\tilde{W_0}$
   \STATE $A_0 \leftarrow \tilde{A_0}[:,0:R_{\text{NMF}}]$
   \STATE $W_0 \leftarrow \tilde{W_0}[0:R_{\text{NMF}},:]$
   \ENDIF
\end{algorithmic}
\end{algorithm}

\subsection{Learning weights $w_{1:M}$ given parameters $\theta_{1:M}$}

To infer the weights corresponding to a given factorization collection $\theta_{1:M}$, we minimize the Stein discrepancy (Algorithm~\ref{alg:eval_framework}) subject to the simplex constraint on the weights. This process involves first computing the pairwise kernel matrix\footnote{As the kernel $\mathcal{K}_p$ is  positive definite, $\mathbf{K}$  is also positive definite.} $\mathbf{K}$ using the kernel $\mathcal{K}_p$ in equation~\ref{eqn:stein_kernel}. The objective function is convex and can be solved using standard convex optimization solvers. Given point-masses $\theta_{1:M}$, this framework can be employed to infer weights for discrete approximations to any posterior for which the score function $\nabla_{\theta} \log p(\theta)$ can be computed.

\begin{algorithm}[t]
   \caption{Kernalized Stein inference for discrete approximations to posterior}
   \label{alg:eval_framework}   
\begin{algorithmic}
\STATE {\bfseries Input:} Particles $\theta_{1:M}$, Score $\nabla_{\theta} \log p(\theta)$, RKHS $\mathcal{H}$ defined by kernel $k$
   \STATE {\bfseries Step 1:} Compute pairwise kernel matrix $\mathbf{K_{i,j}} = \mathcal{K}_p(\theta_i,\theta_j)$ (from equation~\ref{eqn:stein_kernel})
   \STATE {\bfseries Step 2:} Find probability masses that minimize the Stein discrepancy for the given point-masses: $\mathbf{w^*} = \arg \min_{\mathbf{w}} \mathbf{w}^T\mathbf{K}\mathbf{w} \quad \text{s.t} \quad \mathbf{w} \in \Delta^{M-1}$ via standard convex optimization. 
    \STATE {\bfseries Output:} Probability masses $\mathbf{w^*}$
 \end{algorithmic}
\end{algorithm}

\section{Threshold-based Bayesian NMF Model}
The procedure described in Algorithm~\ref{alg:posterior_Q} for finding a discrete approximation to the Bayesian NMF posterior does not depend on any special properties (such as conjugacy) and only requires the joint density $p(X,W,A)$ to be differentiable in order to make inference tractable. In the following, we take advantage of this added freedom to consider a class of models that correspond to what practitioners are directly concerned with. The purpose of this class of models is to allow practitioners to take any application-specific notion of a high-quality factorization and put it into a Bayesian context. 

The ability to incorporate application-specific notions of factorization quality is valuable because notions vary. Squared Euclidean distance is used in hyperspectral unmixing \citep{bioucas2012hyperspectral}, Kullback-Leibler divergence in image analysis \citep{lee2001algorithms} and Itakura-Saito divergence for music analysis \citep{fevotte2009nonnegative}. \citet{roberts2016navigating} notes that these objectives typically only approximate factorization quality, and so, practitioners may be equally interested in all factorizations that meet some quality criteria defined by the the objective. In the Bayesian NMF setting, we would like to construct likelihood models that reflect these application-specific preferences of practitioners. 
 
\subsection{Likelihood: Soft Insensitive Loss Function (SILF) over NMF objectives}
\begin{figure}[t]
\centering
\includegraphics[width=0.4\textwidth]{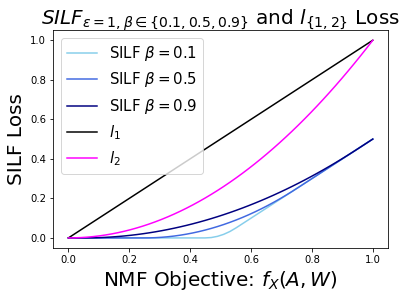}
\caption{A comparison of SILF loss and commonly used $l_1, l_2$ loss functions. The SILF insensitivity parameter $\epsilon$ is set to 0.5, and the smooth transition factor $\beta$ is varied. Small values of $\beta$ lead to sharp transition in the SILF loss profile whereas the transition is less abrupt for large values of $\beta$. In contrast, other popular loss functions such as $l_1$ or $l_2$ do not have insensitive regions, and in the case of NMF, treat the objective function as the sole guide for factorization quality.}
\label{fig:silf}
\end{figure}

We define a likelihood that is maximum (and flat) in the region of high quality factorizations and decays as factorization quality decreases. To do so, we use the soft insensitive loss function (SILF) \citep{chu2004bayesian}: a loss function defined over the real numbers $\mathbb{R}$, where the loss is negligible in some region around zero defined by the insensitivity threshold $\epsilon$, and grows linearly outside that region (see figure~\ref{fig:silf}). A quadratic term depending on the smoothness parameter $\beta$, makes the transition between the two main regions smooth. This transition region has length $2\beta$, making smaller values of $\beta$ correspond to sharper transitions between the flat and linear loss regions. We adapt the SILF from \citep{chu2004bayesian} to only be defined over the non-negative numbers $\mathbb{R}_+$ (as is typical with NMF objectives) and define it as:

\begin{equation*}
\text{SILF}_{\epsilon, \beta}(y) = \left\{
        \begin{array}{ll}
            0 & \quad 0 \leq y \leq (1-\beta)\epsilon \\
            \frac{(y - (1 - \beta))^2}{4\beta\epsilon} & \quad (1 - \beta)\epsilon \leq y \leq  (1 + \beta)\epsilon \\
            y - \epsilon & \quad y \geq (1 + \beta)\epsilon 
        \end{array}
    \right.
\end{equation*}

To form the likelihood, we apply the SILF loss to an NMF objective $f_X(A,W)$ to give:
\begin{equation}
\label{eqn:likelihood}
P(X | W,A) = \frac{1}{Z} e^{(-C \times \text{SILF}_{\epsilon, \beta}(f_X(A,W)))}
\end{equation}

We emphasize that the SILF-based likelihood allows the practitioner to use an NMF objective $f_X(A,W)$ that is best suited to their task and can specify a threshold under that objective for identifying high-quality factorizations.

\paragraph{Prior: Uniform over basis and unambiguous in factorization scaling}
Prior distributions can make the Bayesian NMF problem more identifiable (and possibly unimodal). This is undesirable if the prior is chosen for computational convenience while the landscape of the likelihood has multiple non-trivially distinct modes and connected regions of NMF factorizations (e.g. synthetic data in \citet{Laurberg}). In order to avoid unintentionally altering useful properties of the likelihood landscape in the posterior, we will use a prior that is uniform over the basis matrix $A$. 

NMF has a scaling and permutation ambiguity (equation~\ref{eqn:ambiguity}) that is uninteresting in practice, but depending on the priors chosen, can add redundancy to the posterior distribution. To facilitate exploration of the space of high-quality factorizations, we design our discrete Bayesian NMF posteriors so that meaningful uncertainty in factorization parameters is captured with the least amount of redundancy. We do this by defining the support of our NMF prior on a manifold that eliminates redundancy due to scale.   

\begin{multline}
AW = \underbrace{ASP}_{\widetilde{A}}\underbrace{(SP)^{-1}W}_{\widetilde{W}}  \enskip \text{where S is a positive diagonal matrix, P is a permutation matrix} 
\label{eqn:ambiguity}
\end{multline}

A natural choice of a prior that meets the uniformness and scaling criteria is to have each column of the basis matrix $A_r$ be generated by a symmetric Dirichlet distribution with parameter $\alpha = 1$. This prior determines a unique scale of the factorization and is uniform over the basis matrix $A$ for that scaling. For $W$, we use a prior where each entry $W_{r,n}$ is i.i.d from an exponential distribution with parameter $\lambda_{r,n}$. The exponential distribution has support over all $R_{+}$ ensuring that any weights matrix $W$ corresponding to a column-stochastic basis matrix $A$ is a valid parameter setting under our model, and that the posterior is proper. 

\begin{equation*}
p(A) = \prod_{r = 1}^R p(A_r), \quad A_r \sim \text{Dir}(\mathbf{1}_D)
\end{equation*}

\begin{equation*}
p(W) =  \prod_{n = 1}^N\prod_{r = 1}^R p(W_{r,n}), \quad W_{r,n} \sim \text{Exp}(\lambda_{r,n})
\end{equation*}

\section{Experimental Setup}
In this section, we provide details of our experimental settings and parameter choices, and describe our baseline algorithms and datasets. Our experiments are performed on many benchmark NMF datasets as well as on Electronic Health Records (EHR) data of patients with Autism Spectrum Disorder (ASD) that is of interest to the medical community (see quantitative and qualitative results in Section~\ref{sec:results}).

\subsection{Experimental Settings}

\paragraph{Model Parameters:} While any objective can be put into the SILF likelihood, in the following, we used the squared Frobenius objective $f_X(A,W) = \| X - AW \|_F^2$. To set the threshold parameter $\epsilon$ for each dataset, we use an empirical approach where we find a collection of $50$ high-quality factorizations under default settings of scikit-learn \citep{pedregosa2011scikit}. The objective function is evaluated for each of them $\{f_i\}_{i = 1}^{50}$ and $\epsilon = 1.2 \max_i {f_i}$. We set the remaining SILF likelihood sensitivity parameters $\beta = 0.1$, $C = 2$. For the prior, we identically set the exponential parameter for each entry: $\lambda_{r,n} = 1$.  
 
\paragraph{Stein Discrepancy Base RKHS:} The Stein discrepancy for our variational objective requires a function space to optimize over. This optimization over the function space has an analytical solution when a Reproducing Kernel Hilbert Space (RKHS) is used. \citet{gorham2017measuring} show that the Inverse Multiquadric (IMQ) kernel is a suitable kernel choice for Stein discrepancy calculations as it detects non-convergence to posterior\footnote{\citet{gorham2017measuring} prove that popular Gaussian and Matern kernels fail to detect non-convergence when the dimensionality of its inputs is greater
than 3.} for $c > 0$ and $b \in (-1,0)$.

\begin{equation*}
k_{\text{IMQ}}(\theta_i,\theta_j) = (\|\theta_i - \theta_j\|^2 + c^2)^b
\end{equation*}

Since the length scales of the basis and weights matrix differ, we define a kernel via a linear combination of two IMQ kernels defined separately over the basis $A$ and weights $W$.  

\begin{equation}
k([A_1,W_1],[A_2,W_2]) = \frac{1}{2\gamma_A}(\|A_1 - A_2\|^2 + c_A^2)^{b_A} + \frac{1}{2\gamma_W}(\|W_1 - W_2\|^2 + c_W^2)^{b_W} 
\label{eqn:my_imq}
\end{equation}
Here $\gamma_A = (c_A^2)^{b_A}$ and similarly $\gamma_W = (c_W^2)^{b_W}$ are scaling factors that ensure the kernel takes values between 0 and 1. In general, across our datasets, the Dirichlet prior on the basis matrix induces a small length scale for $A$ and a larger length scale for the weights $W$. We uniformly set $c_A = 1 \times 10^{-2}$, $c_W = 1 \times 10^{3}$ and $b_A =  b_W = -0.5$ across all our datasets. While these parameters can be chosen to be data-specific, our kernel similarity analysis shows agreement with pairwise distances between basis and weights matrices of obtained factorizations (see figures~\ref{fig:kernel},~\ref{fig:basis},~\ref{fig:weights} in supplementary materials), indicating that the parameter choice is reasonable.  

\paragraph{Software for inferring weights $w_{1:M}$:}
The optimization for the weights $w_{1:M}$ (Step 2 in Algorithm~\ref{alg:eval_framework}) is carried out using the Splitting Conic Solver (SCS) in the convex optimization package CVXPY~\citep{diamond2016cvxpy}. 

\paragraph{Transfer initialization settings:} For the $Q$-Transform initializations, we set the transfer rank and SVD rank $R_{T} = R_{\text{SVD}} = 3$. We generated twenty sets of synthetic data $X_s \in \mathbb{R}_+^{12 \times 12}$ using non-negative matrices of rank $R_T$ with truncated gaussian noise. For each synthetic dataset, we find five pairs of transformation matrices through random restarts. In all our experiments, the same set of $M_\text{max} = 100$ pairs of transformation matrices $\{Q^m_A,Q^m_W\}_{m =1}^{100}$ are applied to each of the real datasets.

\subsection{Baselines}
Our transfer learning approach provides an indirect way of creating a posterior distribution. We find candidate factorizations through initializing NMF algorithms with $Q$-Transform initializations and incorporate them into a discrete approximation of the Bayesian NMF posterior. The posterior approximations are evaluated using the Stein discrepancy which is a principled variational objective. In this section, we compare our procedure with other initialization strategies, an MCMC baseline, and two gradient-based approaches that directly incorporate the Stein discrepancy kernel and Bayesian NMF model in their gradient updates. 

\begin{itemize}
\item \textbf{Initialization Approaches} involve running our main algorithm for particle-based inference for Bayesian NMF with the modification that we replace the $Q$-Transform initialization (Step 2 of  Algorithm~\ref{alg:posterior_Q}) with the alternate ways of initialization. 
In experiments, we keep track of the time taken (initialization and optimization) to produce each of the $M_{\text{max}} = 100$ factorizations. By sampling collections of size $M < M_\text{max}$ from this collection and computing Stein Discrepancies for the approximate NMF posteriors, we compare the performance of these initialization approaches in terms of Stein discrepancy and runtime.  
\begin{itemize}

\item \textbf{Random} restart initializations for NMF in scikit-learn \citep{pedregosa2011scikit} set each entry of the factors $A,W$ as independent, coming from a truncated standard normal distribution. These entries are all scaled by $\eta = \sqrt{\frac{1}{R_{\text{NMF}}}\sum_{D,N} X_{d,n}}$ and are given by: $A_{d,k}^0 , W_{k,n}^0 \sim  \eta | \mathcal{N}(0, 1) | $.
\item  \textbf{NNDSVDar} initialization is a variant of a popular initialization technique called Nonnegative Double Singular Value Decomposition (NNDSVD) which was introduced by \citet{boutsidis2008svd}. It is based on approximating the SVD expansion with non-negative matrices. Since the NNDSVD algorithm is deterministic, this only gives a single initialization. The NNDSVDar variant of this initialization replaces the zeros in the NNDSVD initialization with small random values. We use the scikit-learn initialization for NNDSVDar which uses a randomized SVD algorithm \citep{halko2009finding}, and note that it introduces some additional variability in the initializations. \end{itemize}
\item\textbf{Markov Chain Monte Carlo} approaches involve sampling from a Markov Chain whose stationary distribution is the posterior of interest. In this work, we use Hamiltonian Monte Carlo (HMC). \textbf{HMC} is initialized with an NMF obtained using the default settings of scikit-learn \citep{pedregosa2011scikit} (warm start), and adaptively selects the step-size using the procedure outlined in \citet{Neal}. Since our prior on $A$ restricts the columns of $A$ to belong to the simplex, and the entries of the weights matrix to all be non-negative, we need to simulate Hamiltonian dynamics as defined on the manifold that forms the support of the posterior. To do this, we incorporate a reparametrization trick \citep{betancourt2012cruising, altmann2014unsupervised} to sample under the column-stochastic (simplex) constraints of the basis matrix $A$, and a mirroring trick \citep{patterson2013stochastic} for sampling from the positive orthant for the weights matrix $W$. We run the chain for 10000 samples and then thin it to $M_{\text{max}} = 100$ factorizations and compute the Stein discrepancy using Algorithm~\ref{alg:eval_framework} with $M$ factorizations where $M < M_{\text{max}}$. We repeat this experiment three times to capture variability in the performance of the HMC. 
\item \textbf{Gradient Approaches} are designed to optimize via gradient descent, a collection of factorizations so that they can represent the Bayesian NMF posterior. This approach requires fixing the size of the collection. In our experiments, we set the size of this collection to be equal to $M = 5$. Due to the large memory requirement of running this algorithm with automatic differentiation using autograd \citep{maclaurin2015autograd}, we were unable to run these algorithms for larger $M$. We impose scaling and non-negativity constraints after every gradient step (for a total of 1000 steps) and keep track of the Stein discrepancy in relation to the algorithm's runtime. The experiment is repeated three times to capture variability in its performance over multiple iterations. We use the following two algorithms:
\begin{itemize}
\item \textbf{SVGD}: Stein Variational Gradient Descent is a functional gradient descent algorithm \citep{liu2016stein} that optimizes a collection of particles (factorizations) to approximate the posterior. We replace the RBF kernel from the original work with the IMQ-based kernel defined in equation~\ref{eqn:my_imq}. 
\item  \textbf{DSGD}: Direct Stein Gradient Descent. We replace the functional gradient descent of SVGD with the gradient of the Stein discrepancy (using automatic differentiation \citep{baydin2015automatic} for python code with autograd \citep{maclaurin2015autograd}). 
\end{itemize}

\end{itemize}

\subsection{Datasets}

Our datasets cover a range of different types and can be divided into three main categories (count/rating data, greyscale face images and hyperspectral images). The ranks for hyperspectral data are chosen according to ground truth values. In the 20-Newsgroups data, we select articles from 16 newsgroups (hence the rank 16) and for other datasets we pick a rank that corresponds to explaining at least 70 percent of the variance in the data (as measured by the SVD). Table \ref{table:data_sets} provides a description of each dataset as well as the rank used and a citation. The Autism dataset is of interest to the medical community for understanding disease subtypes in the Autism spectrum and is not publicly available. The remaining datasets are public and are considered standard benchmark datasets for NMF.

\begin{table}[!htbp]
\centering
\caption{Datasets for NMF}
\label{table:data_sets}
\begin{tabular}{| l | c|  c| c | r |  }
\hline
Dataset                & Dimension & Observations & Rank  & Description \\ \hline
20-Newsgroups          &   1000 & 8926                &   16       &  Newspaper articles \citep{562The2065:online} \\

Autism                       &    2862 & 5848               &   20          &  Patient visits \citep{doshi2014comorbidity}          \\ 
 
\hline
 LFW     & 1850 & 1288      & 10    &  Grayscale Faces Images \citep{LFWFaceD94:online}      \\

Olivetti Faces         & 4096 &  400       & 10      &   Grayscale Faces Images \citep{FacesData}        \\
Faces  CBCL        & 361 & 2429       & 10      &  Grayscale Faces Images \citep{HomePogg44:online}   \\ 

Faces  BIO            & 6816 & 1514      & 10      &  Grayscale Faces Images \citep{BioIDFac89:online}    \\ \hline
Hubble                 & 100 & 2046       & 8   &   Hyperspectral Image \citep{HubbleData}     \\
Salinas A               & 204 & 7138       & 6      &  Hyperspectral Image \citep{MultiSpe80:online}  \\
Urban                  & 162 & 10404      & 6     &   Hyperspectral Image  \citep{fyzhu_2014_IJPRS_SS_NMF}   \\ \hline
\end{tabular}
\end{table}

\section{Results}
In this section, we compare computational time and Stein discrepancy values for variational posteriors obtained through different algorithms. We find that our approach for Bayesian NMF posterior approximation using transfer learning ($Q$-Transform) consistently produces the highest quality posterior approximations in the shortest amount of time. Inspection of factorization parameters from $Q$-Transform reveals that the parameter uncertainty captured by the Bayesian NMF posterior approximation has meaningful consequences for interpreting and utilizing these factorizations.

\label{sec:results}
\subsection{Quantitative Evaluation of Posterior Quality}
\label{sec:quantitative}
\begin{figure}[!htbp]
\centering
\includegraphics[width=.3\textwidth]{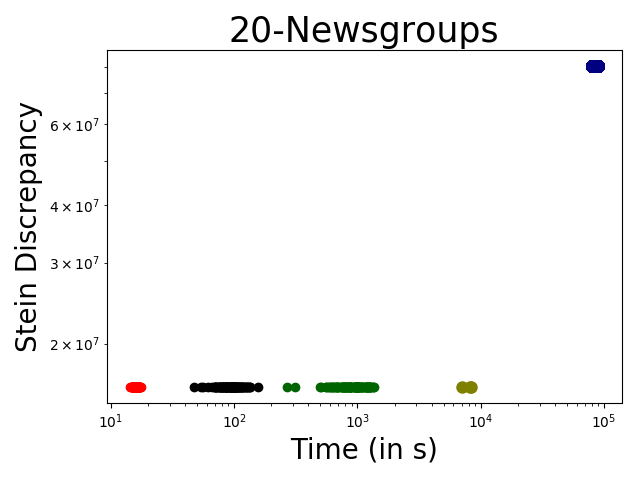}\quad
\includegraphics[width=.3\textwidth]{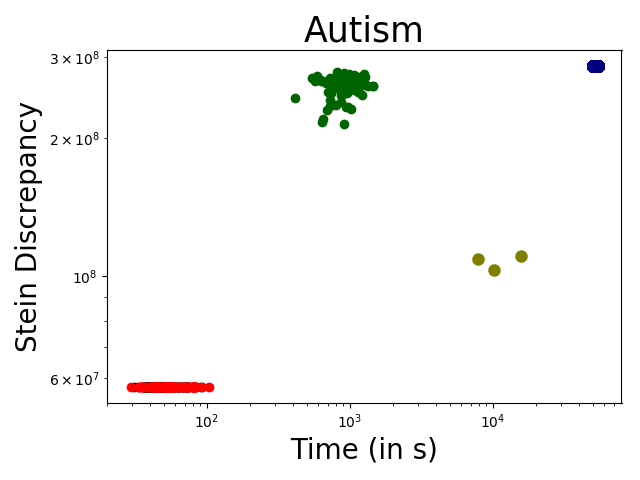}\quad
\includegraphics[width=.3\textwidth]{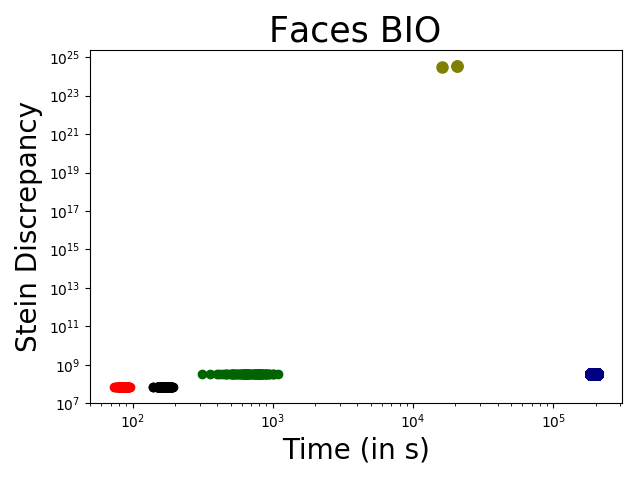}

\medskip

\includegraphics[width=.3\textwidth]{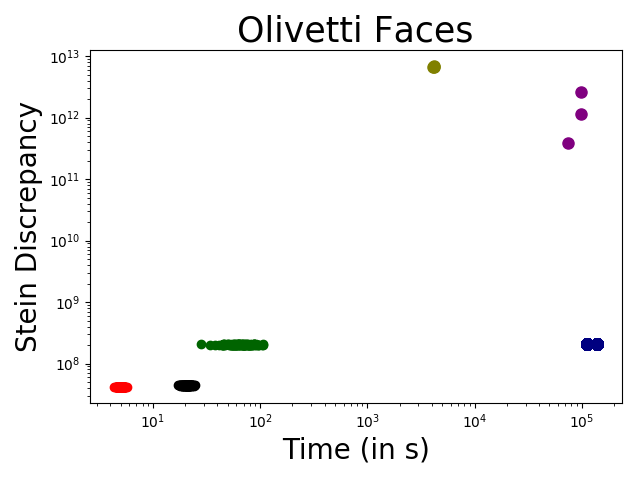}\quad
\includegraphics[width=.3\textwidth]{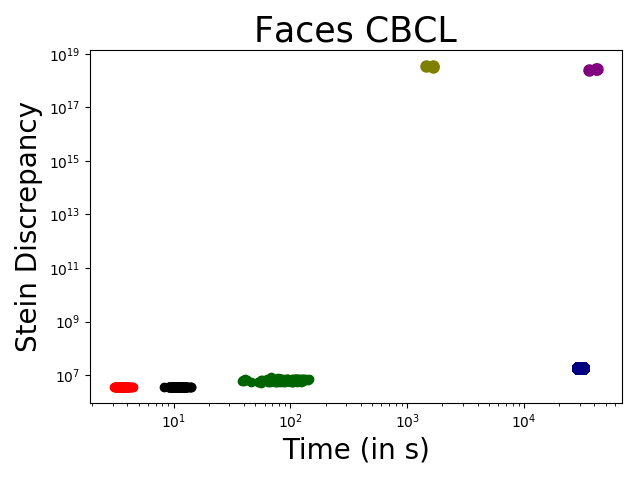}\quad
\includegraphics[width=.3\textwidth]{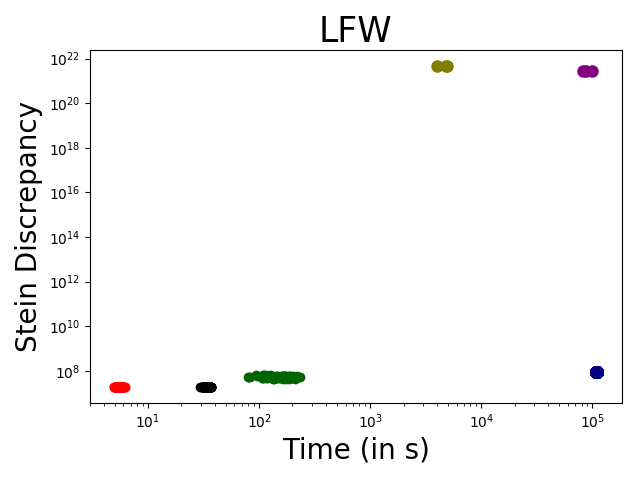}

\medskip 

\includegraphics[width=.3\textwidth]{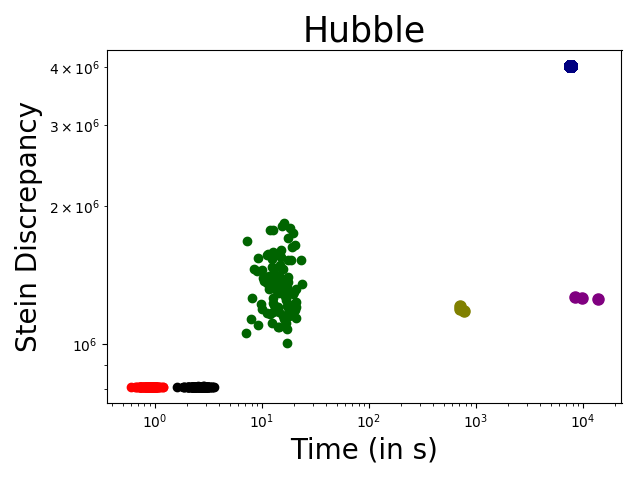}\quad
\includegraphics[width=.3\textwidth]{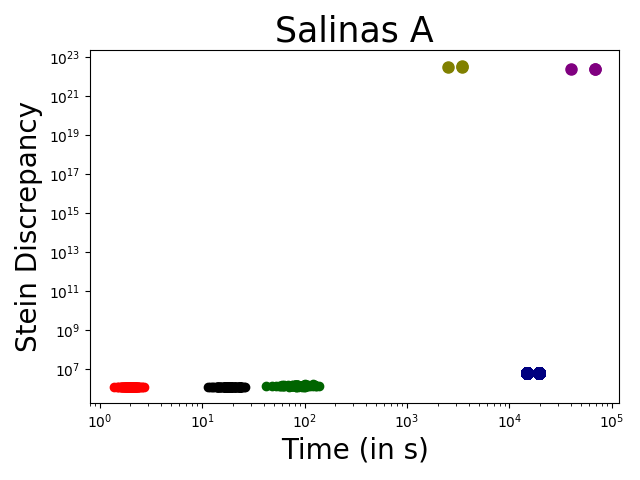}\quad
\includegraphics[width=.3\textwidth]{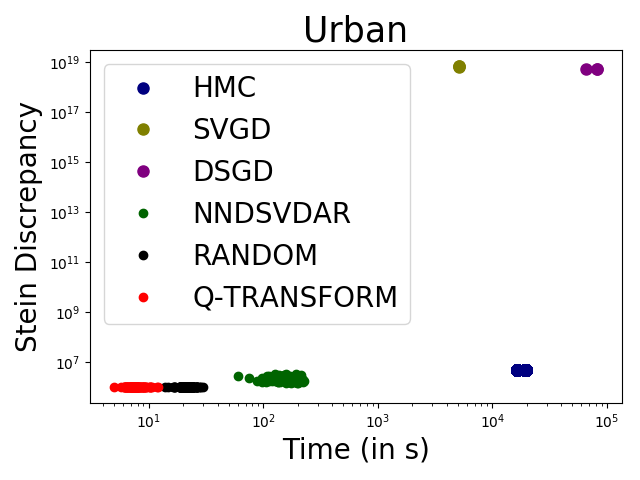}

\caption{For each dataset we show the the quality of the Bayesian NMF approximate posterior ($M = 5$) and the corresponding runtime of $Q$-Transform and the other baselines. Across multiple datasets, we see that the best discrete posteriors to Bayesian NMF (lowest Stein discrepancy) are produced in the least time using the Q-Transform initializations (in red).}
\label{fig:qr}
\end{figure} 
 
The Stein discrepancy variational objective involves terms that consider both the quality of the factorizations (as given by the score function $\nabla_\theta \log p(\theta)$ ) and their similarity (as given by the base RKHS kernel $k(\theta_i,\theta_j)$). The initialization-based approaches ($Q$-Transform, Random and NNDSVD) are all designed to find high quality factorizations (as defined by the SILF Likelihood). The diversity of high-quality factorizations therefore plays an important role in determining the Stein discrepancy. In figure~\ref{fig:qr}, we see that both $Q$-Transform and the approach that substitutes $Q$-Transform initializations for random restarts (labelled Random), obtain the highest quality posterior approximations for $M = 5$ and the $Q$-Transform approach consistently takes less time. The NNDSVDar initializations and thinned HMC samples lead to factorizations that are high-quality but often not diverse (see diversity analysis in supplementary material: figures~\ref{fig:kernel},~\ref{fig:basis},~\ref{fig:weights}). The SVGD and the DSGD are generally the worst performing algorithms. These methods are often unable to find factorization parameters that meet the quality criteria of the SILF likelihood (see quality analysis in supplementary material: figures~\ref{fig:llik} and~\ref{fig:fro}). This is understandable because even using simple gradient-based approaches to find a single high-quality NMF turns out to be difficult, hence the existence of a literature on specialized algorithms for performing NMF.

\begin{figure}
\centering
\includegraphics[width=.3\textwidth]{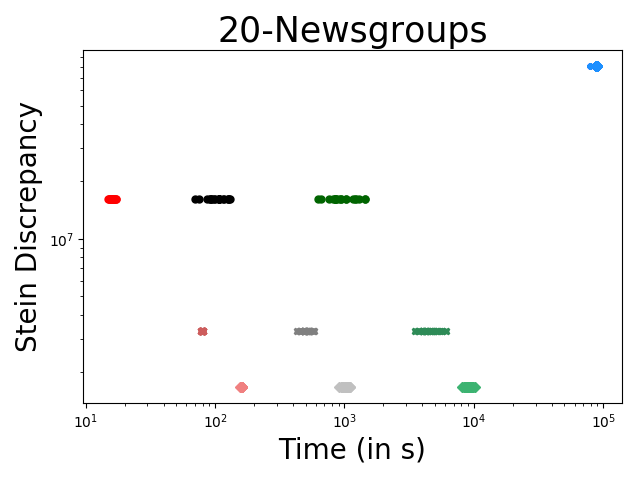}\quad
\includegraphics[width=.3\textwidth]{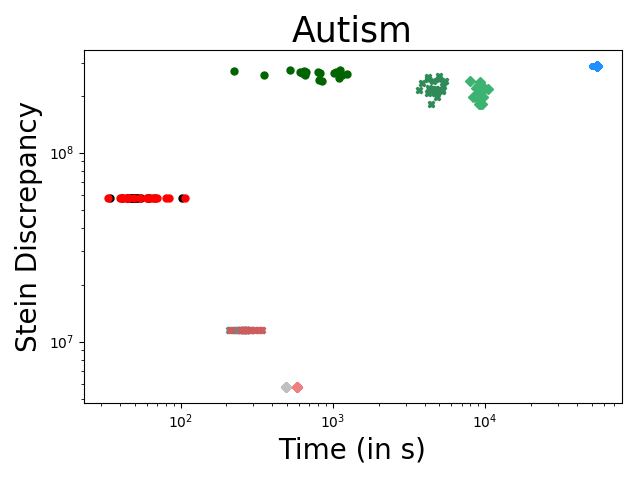}\quad
\includegraphics[width=.3\textwidth]{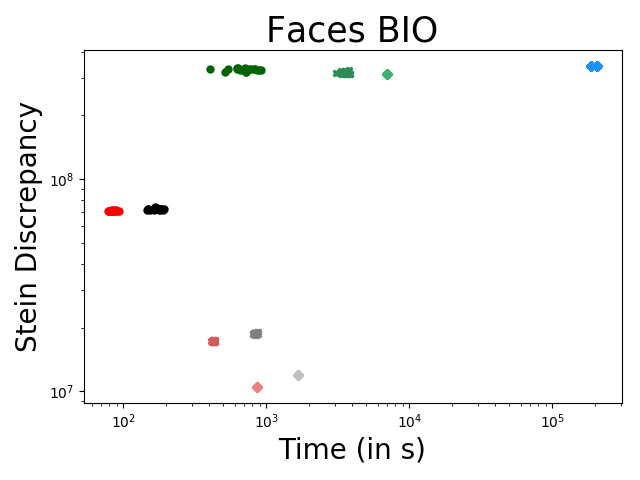}

\medskip

\includegraphics[width=.3\textwidth]{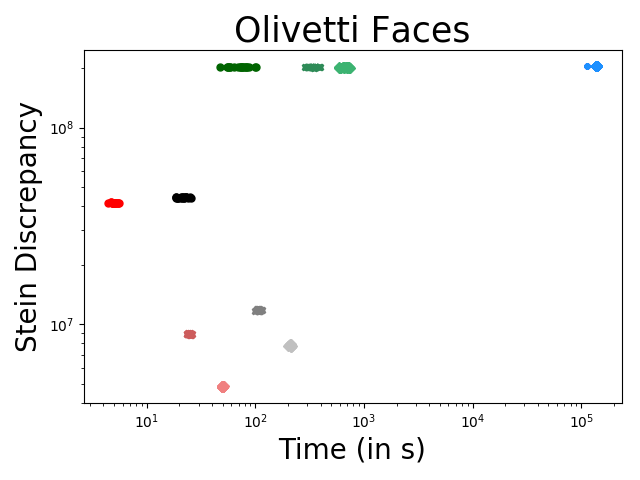}\quad
\includegraphics[width=.3\textwidth]{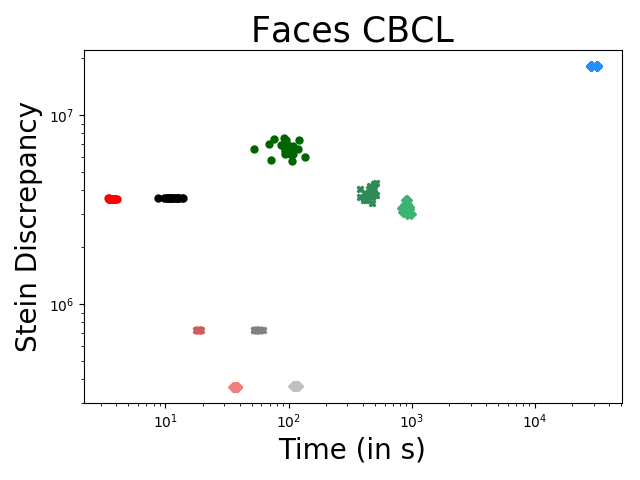}\quad
\includegraphics[width=.3\textwidth]{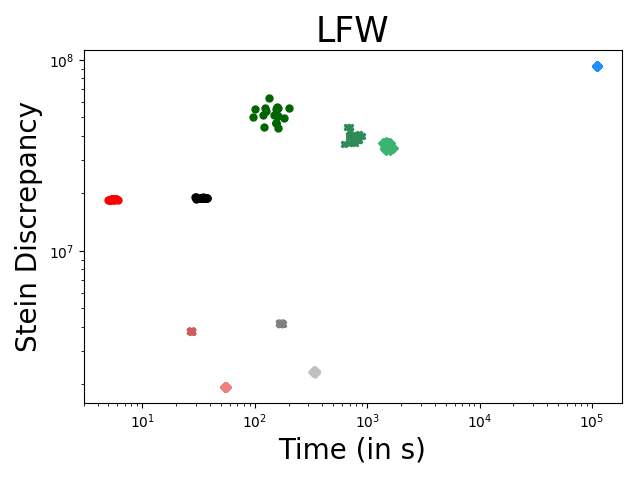}

\medskip 

\includegraphics[width=.3\textwidth]{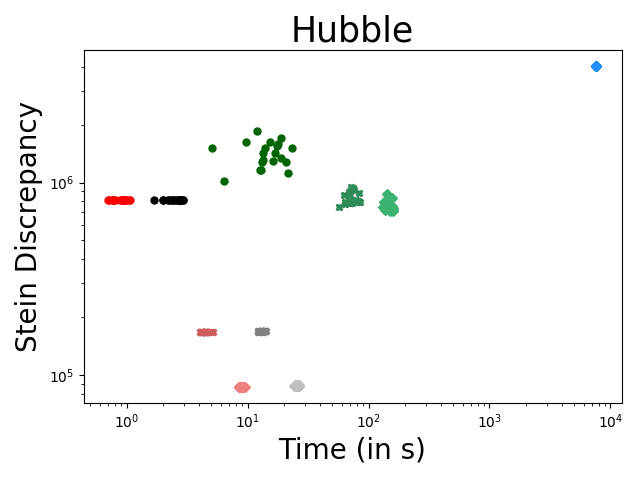}\quad
\includegraphics[width=.3\textwidth]{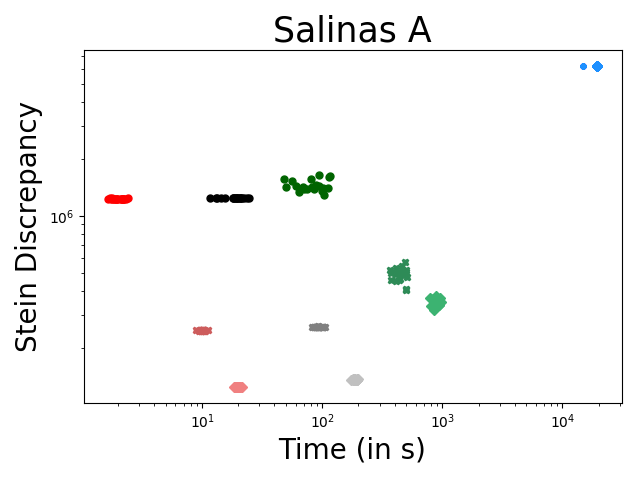}\quad
\includegraphics[width=.3\textwidth]{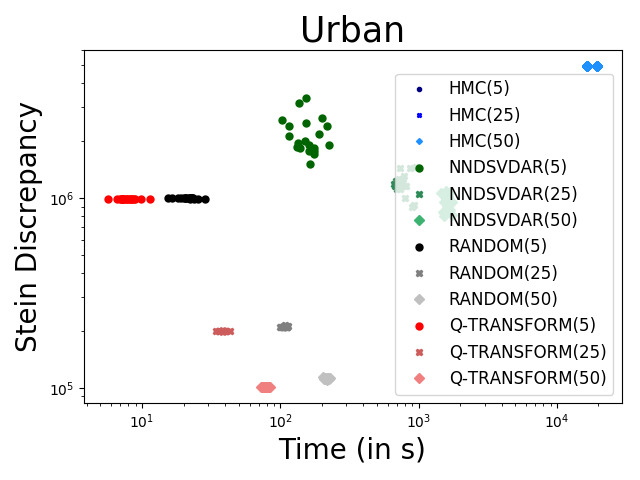}

\caption{For each dataset we show the the quality of the Bayesian NMF approximate posterior (for $M = \{5,25,50\}$) and the corresponding runtime of $Q$-Transform and the other baselines (except SVGD and SDGD). Across multiple datasets, we see that the best discrete posteriors to Bayesian NMF (lowest Stein discrepancy) are produced in the least time using the Q-Transform initializations (in red).}
\label{fig:qr_multi}
\end{figure}

Figure~\ref{fig:qr_multi} shows results for $M = \{5, 25, 50\}$ where $Q$-Transform continues to have a runtime advantage over other baselines. Additionally, for some datasets (Olivetti Faces, LFW and Faces BIO) $Q$-Transform also produces higher quality of the posterior approximations. Variational posteriors constructed using thinned samples from HMC significantly lack diversity as the Stein discrepancies for collections of size 5, 25 and 50 are comparable. This indicates that the HMC chain only explores a small region of the posterior distribution and can be confirmed through the diversity analysis in the supplementary material (figures~\ref{fig:kernel},~\ref{fig:basis},~\ref{fig:weights}). \citet{sminchisescu2007generalized} notes that in high dimensional spaces, we expect there to be many ridges of probability as there are likely to be some directions in which the posterior density decays sharply. Alternatively, there may be several isolated modes with no connecting regions of high probability making it particularly challenging for the HMC  chain to avoid getting stuck in a local mode of the Bayesian NMF posterior.

\subsection{Interpretation and Utilization of Posterior Estimates}
NMF posteriors can provide insight into the non-identifiability present within a particular dataset. Different factorizations may explain the data as a whole equally well, but do it through dictionary elements that have different interpretations or can be used to understand specific parts of the data better than other factorizations. We show visual examples of diversity in the top words of the 20 Newsgroups NMF posteriors and examples of how performance in downstream tasks for the 20 Newsgroups and Autism dataset is dependent on the posterior samples. Our analysis yields meaningful insights that could not be gained through a single factorization.

\begin{figure}
\centering
  \includegraphics[width=0.75\textwidth]{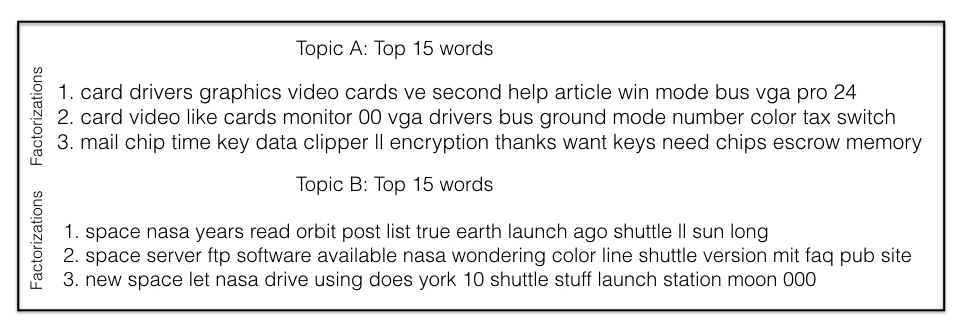}
  \caption{The top 15 words for topic A (computers/electronics) and topic B (space) shows that different factorizations provide an emphasis on different terms. In topic A, the top word from factorization 1 and 2 is `card', but it does not appear in the top 15 words of factorizations 3. Instead a similar term `chip' is emphasized in Factorization 3. In topic B, the terms `space' and `nasa' appear in all three factorizations but factorization 2 is the only one with digital terms like `ftp', 'server','site' and 'faq'. In contrast factorization 1 and 3 both contain more physical terms like `sun', `moon',`launch'. }
  \label{fig:qualitative2_20ng}
\end{figure}

\paragraph{20-Newsgroups}

\begin{figure}
\centering
\includegraphics[width=0.5\textwidth]{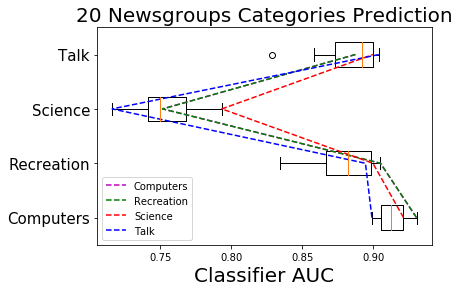}
\caption{Classifiers trained on feature vectors from different factorizations yield variability in prediction performance (as measured by AUC). The dotted lines show the factorization that produces the best performing classifier for each category. The factorization (blue dotted line) that predicts the `Talk' category best is actually one of the worst performing factorizations for the `Science' category. This variability in performance demonstrates that no single factorization gives the best latent representation for the overall prediction task.}
\label{fig:quantitative_20ng}
\end{figure}

Our Bayesian NMF of 20-Newsgroups in section~\ref{sec:quantitative} was a rank 16 decomposition of posts from 4 categories.  In figure~\ref{fig:quantitative_20ng}, we show the held-out AUC of a classifier trained to predict those categories based on the weight matrix $W$ from each factorization in our variational posterior.  Even though all of these factorizations have essentially equivalent reconstruction (see figure~\ref{fig:fro} in supplementary material), there exists a significant variation in the performance of these NMFs on the prediction tasks. The best performing NMF for one category is generally not the best (or even one of the top performing) NMFs for other categories.  This observation may be valuable to a practitioner intending to use the NMF for some downstream task: different samples explain different patterns in the data.  In figure~\ref{fig:qualitative2_20ng}, we see that this is indeed true: even after alignment,\footnote{We compare topics after finding the permutation of columns that best aligns them by solving the bipartite graph matching problem. We minimize the cost given by the angle between topics.} distinct NMF factorizations have top words that indicate different emphasis across topics.

\paragraph{Autism Spectrum Disorder (ASD)}

\begin{figure}
\centering 
  \begin{subfigure}{.43\linewidth}
    \includegraphics[width=\linewidth]{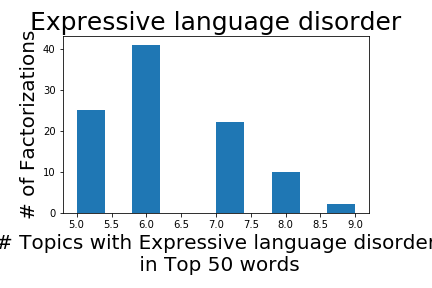}
  \end{subfigure}
  \begin{subfigure}{.43\linewidth}
    \includegraphics[width=\linewidth]{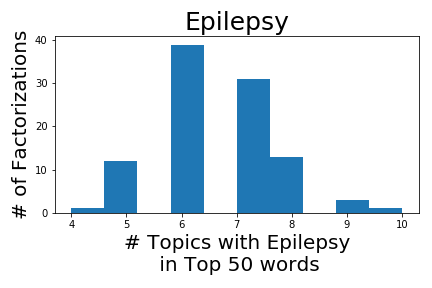}
  \end{subfigure}  
  \begin{subfigure}{.43\linewidth}
    \includegraphics[width=\linewidth]{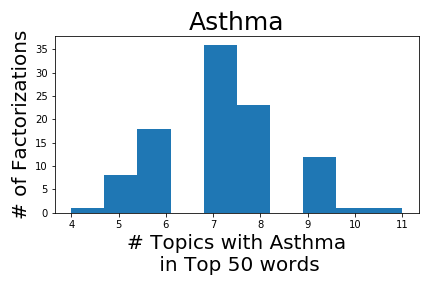}
  \end{subfigure}
  \begin{subfigure}{.43\linewidth}
    \includegraphics[width=\linewidth]{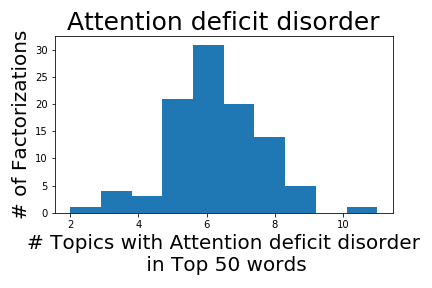}
  \end{subfigure}
  \caption{We explore top words in the topics relating to key terms of interest to clinicians and discover that different NMFs place varying amount of emphasis on different terms. Such variability is of interest to clinicians who may be trying to interpret topics to understand patterns in ASD.}
  \label{fig:qualitative_asd} 
\end{figure}

In addition to core autism symptoms, \citet{doshi2014comorbidity}
describe three major subtypes in autism spectrum disorder: those with
higher rates of neurological disorder, those with higher rates of
autoimmune disorders, and those with higher rates of psychiatric
disorders.  In figure~\ref{fig:qualitative_asd}, we show the number of
topics that contain key terms corresponding to these areas
(expressive language disorder, epilepsy, asthma, and attention deficit
disorder) across different factorizations in the variational posterior obtained via $Q$-Transform.  The large variation
suggests that different factorizations in the particle-based posterior are spending different
amount of modeling effort across these known factors; knowing that
such uncertainty exists is essential for clinicians who may be trying
to interpret topics to understand patterns in autism spectrum disorder.

 \begin{figure}
\centering
\includegraphics[width=0.45\textwidth]{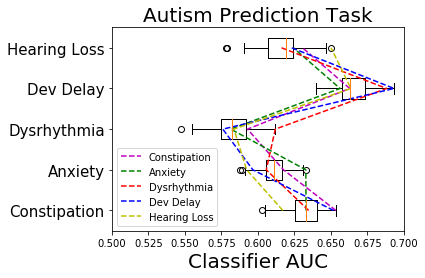}
\caption{Classifiers trained on weights matrix of $M_{\text{max}} = 100$ different factorizations to predict the presence of certain medical codes in a patient's trajectory exhibit significant variability in prediction on a test set (as measured by AUC). Different factorizations lead to top predictors for the onset of different medical issues.}
\label{fig:autism_auc}
\end{figure}

 On the same set of patients, we can also ask whether we can predict the onset of certain medical issues in the subsequent patient trajectory. We train a classifier on the weights of the NMFs to predict the onset of these medical issues. Similar to the category prediction results in 20-Newsgroups, figure ~\ref{fig:autism_auc} shows that there is a large variability (around 0.1 in AUC) in the performance of classifiers trained on the weights matrices of different factorizations on the prediction task. No single factorization has the best performance across the different prediction tasks.

\subsection{Extension: Bayesian NMF in the presence of missing data}
In the presence of missing data, there is perhaps an even greater need to understand the uncertainty in factorization parameters for NMF. The factorization space of a fully observed dataset forms a subset of the factorization space in the presence of missing data. Our particle-based approach to Bayesian NMF posterior approximation can be applied to the missing data setting by making some minor adjustments to the experimental settings.

\begin{figure}[!htbp]
\centering

\includegraphics[width=.3\textwidth]{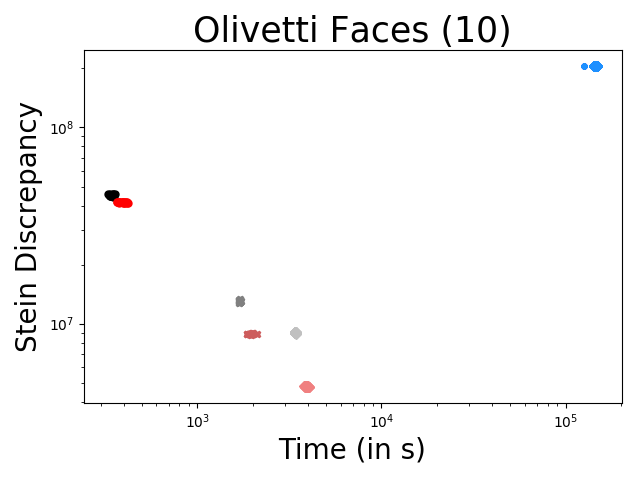}\quad
\includegraphics[width=.3\textwidth]{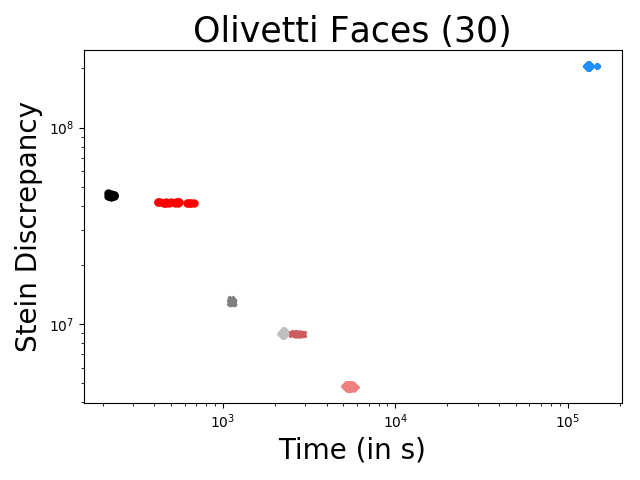}\quad
\includegraphics[width=.3\textwidth]{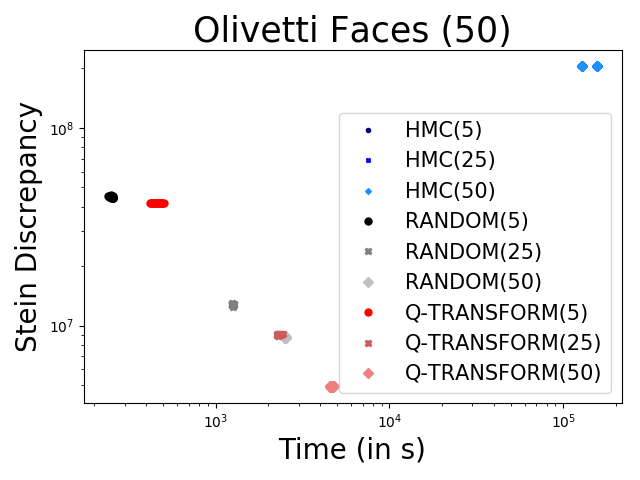}

\caption{Under different percentages of missing-ness in the Olivetti Faces dataset (10\%, 30\%, 50\%), the quality of the Bayesian NMF approximate posterior and the corresponding runtime of $Q$-Transform and the other baselines is shown. The best discrete posterior approximations to Bayesian NMF are produced using the Q-Transform initializations (in red).}
\label{fig:quant_missing}
\end{figure}

The multiplicative update algorithm for NMF \citep{lee2001algorithms} can be adjusted so that the update equations for factorization parameters only consider the observed data. We use an implementation of this modification to the multiplicative update algorithm\footnote{https://github.com/scikit-learn/scikit-learn/pull/8474/commits/a838f94c8c832aaf57140f23bd8c8a14daec2626} to find a completion of the data $X$, compute the SVD subspace and then apply our $Q$-Transform initializations. Figure~\ref{fig:quant_missing} demonstrates that our approach to Bayesian NMF can be extended to the case where the data matrix $X$ is partially observed. For the Olivetti Faces dataset with varying degrees of missing-ness, the $Q$-Transform approach to Bayesian NMF consistently finds posterior approximations  that are significantly better (as measured by Stein Discrepancy) than other baselines whereas for a given $M$, the runtime is second-lowest. 

\begin{figure*}
\centering
  \includegraphics[width=0.85\textwidth]{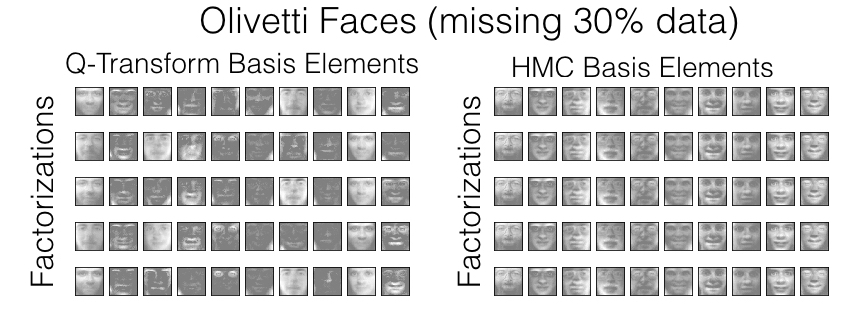}
  \caption{Sample factorizations from the variational posterior using $Q$-Transforms show that a diverse range of basis elements can be use to approximate the data. However, HMC samples seem to be identical indicating that HMC was only exploring a very small region of the posterior space.}
  \label{fig:faces30p}
\end{figure*}

Figure~\ref{fig:faces30p} shows sample factorizations from the variational posterior using $Q$-Transform and HMC samples. To allow for comparison, we have aligned the positions of the basis (dictionary) elements to a reference factorization using the bipartite matching algorithm. It is clear from looking at the $Q$-Transform factorizations that a diverse range of dictionaries can be used to approximate the data well whereas the HMC chain only explores one set of dictionary elements.  Interestingly, the diversity of solutions obtained using $Q$-Transform have visually interpretable differences, i.e. these are not simply perturbations of some ground truth basis elements. Some of the basis elements look like faces and some of them look like different shadow or lighting configurations. In contrast, the factorization samples from HMC have basis elements that look identical. This indicates that the HMC has explored a limited region of the posterior space.

\section{Discussion: When is $Q$-Transform successful?}

Our ability to extract \emph{transferrable} low-rank transformation matrices from an SVD and an instance of NMF indicates that there exist some shared similarities between the SVD and NMFs. In this section we seek to develop a better intuition behind the success of the $Q$-Transform initializations at exploiting these similarities. We investigate the effect of noise on the quality of the initializations and the resulting time-advantage in optimization. In addition, we explore the choice of transfer rank $R_{T}$ and its effect on the quality of initializations. Lastly, we describe a practical implementation consideration regarding consistency of convention used in the SVD.

\subsection{The $Q$-Transform Initialization versus Noise}
\begin{figure}
\centering
\includegraphics[width=.43\textwidth]{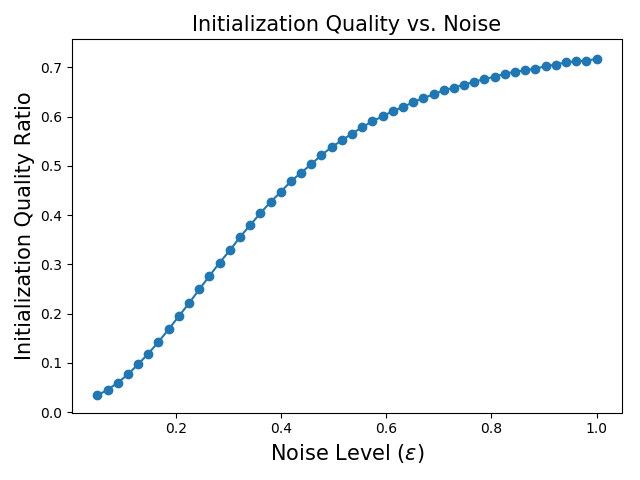}\quad
\includegraphics[width=.43\textwidth]{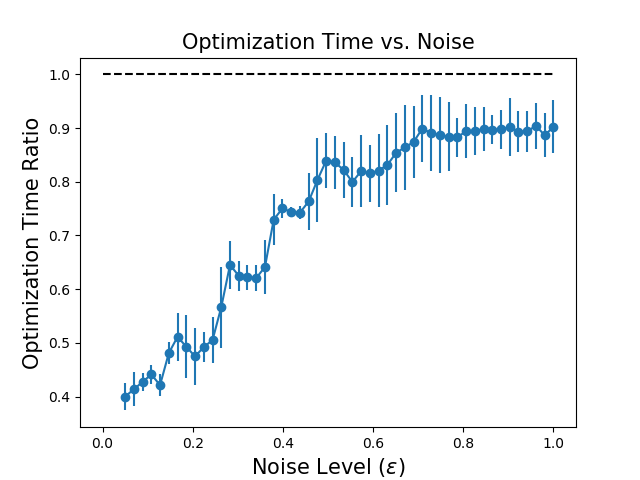}

\caption{In the low-noise regime, the reconstruction error of $Q$-Transform initializations is significantly less than random restart initializations. This relative advantage gets smaller as the noise level increases. Similarly, the time taken to converge is significantly shorter than the random restart approach under the low noise scenario and continues to increase with noise. As expected, at high noise levels there exists no additional advantage to the $Q$-Transform approach (the optimization time ratio approaches 1).}
\label{fig:synthetic}
\end{figure}

In Section~\ref{sec:bnmf_approximation}, we sought high-quality initializations because they generally require less time to converge. On synthetic data $X =  AW + \epsilon N_{o}$  (D = N = 500, K = 20) we explore the effect of increasing noise ($\epsilon$) on the quality of our transfer-based NMF initializations and the time taken to converge. We normalize the norm of the noise matrix to be equal to the norm of the data $\|N_o\| = \|AW\|$ so that the contribution of signal $AW$ and noise $\epsilon N_o$ to the data is equal when $\epsilon = 1$. We continue to use the same $100$ pairs of $Q_A,Q_W$ matrices. We compare the performance of $Q$-Transform over random restarts in terms of initialization quality (ratio of the reconstruction error from $Q$-Transform to the reconstruction error from random restart) and time to convergence (ratio of time taken using $Q$-Transform initialization to time taken using random restart). In both metrics, the $Q$-Transform has an advantage over random restarts for values of the noise $\epsilon$ smaller than $1$, and the advantage is greatest for smallest noise. Figure~\ref{fig:synthetic} shows that the advantage of $Q$-Transform initializations is highest in a low noise regime and decreases as the noise increases. This behavior makes sense because as noise increases, the data is no longer truly low rank.   

\subsection{Selecting transformation dimensions $R_T$ and $R_{\text{SVD}}$:}

Transformation matrices $Q_A, Q_W$ map basis vectors defining the top SVD subspace of dimension $R_{\text{SVD}}$ to a set of $R_{T}$ non-negative basis vectors that approximate the same subspace. The full initialization for NMF is obtained by either padding the initialization with small entries ($R_{T} < R_\text{NMF}$) or removing extra columns and rows of the factor matrices ($R_{T} > R_\text{NMF}$). For simplicity, we consider the case where the transfer rank and SVD rank are equal $R_T = R_{\text{SVD}}$ and the resulting transformation matrices $Q_A, Q_W$ are square. In our experiments on real data, we chose the transfer rank and SVD rank $R_T = R_{\text{SVD}} = 3$ and demonstrate that it successfully produces the highest quality approximate posteriors in the least amount of time. In practice, we could have chosen from a range of values to determine the transfer rank $R_T$ and SVD rank $R_{\text{SVD}}$. 

\begin{figure}
\centering
\includegraphics[width=.5\textwidth]{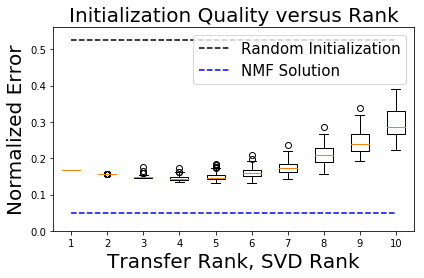}\quad

\caption{On a synthetic dataset ($D = N = 500,  K_{\text{NMF}} = 10$), we apply $Q$-Transform initializations using varying transfer ranks and SVD ranks $R_T = R_{\text{SVD}} = \{1,2,\hdots,10\}$. We see that for a range of low rank values, the $Q$-Transform initializations are high quality, but at larger values the quality of initializations gets worse. The black dotted line shows the highest quality random initialization and the blue dotted line shows the lowest quality NMF solution. The reconstruction errors are normalized by the norm of the data.}
\label{fig:finding_RT}
\end{figure}

To investigate, we extract a set of $100$ transformation matrices $Q_A, Q_W$ for transfer dimensions $R_T = R_{\text{SVD}} = \{1,2,\hdots,10\}$ using synthetic data ($D = N = 15$). Once constructed, we applied the transformation matrices to a $500 \times 500$ dataset $X_{\text{TEST}}$ of rank $K = 10$. Figure~\ref{fig:finding_RT} shows the quality of the initialization from each of the different transfer dimensions explored. Even though the dataset $X_{\text{TEST}}$ has rank $10$, the rank $10$ transformation matrices found using the $15 \times 15$ synthetic dataset are unable to successfully transfer to this new dataset. We see a general trend that as the transfer rank increases, the quality of the initializations gets worse. This tells us that the transformations of the top-most singular vectors of the data most consistently transfer to new datasets whereas the transformations of singular vectors corresponding to smaller singular values are less consistently transferred to new datasets.  

\subsection{Sign Convention for SVD}
In considering when $Q$-Transform is successful, we note that there exists an intrinsic ambiguity in the sign of the singular vectors of $X$: changing the sign of any column of $A_{\text{SVD}}$ and corresponding row of $W_\text{SVD}$ gives a valid SVD. For $Q$-Transform to work, we must apply a consistent resolution of the sign ambiguity (e.g. from \citet{bro2008resolving}). This ensures that learned transformations $Q_A, Q_W$ map in a consistent way to SVD decompositions of new datasets.

\section{Related Work}
\label{sec:related_work}
Recent work on NMF has involved theoretical work on non-identifiability with new algorithms that can provably recover the NMF under certain assumptions \citep{li2017provable, bhattacharya2016non, ge2015intersecting}. However, these assumptions are often difficult to check and may indeed be violated in practice.  

There is a large body of work on Bayesian NMF.  Sampling-based approaches include Gibbs sampling \citep{schmidt2009bayesian}, Hamiltonian Monte Carlo \citep{schmidt2009probabilistic}, and reversible jump variants \citep{schmidt2010reversible}.  All of these have trouble escaping local modes \citep{masood2016empirical}, and are often constrained to a limited class of tractable distributions. Variational approaches to Bayesian NMF have successfully yielded interpretable factorizations \citep{bertin2009fast, cemgil2009bayesian, paisley2014bayesian} but also typically only capture one mode.  We note that in many cases, priors of convenience---for example, exponential distributions---can induce a single dominant mode, even when that was not the intent of the practitioner. \citet{gershman2012nonparametric} develop a non-parametric approach to variational inference that provides flexibility in modeling the number of Gaussian components required to approximate a posterior. However, the isotropic covariance in the model makes it unsuitable for applying it to Bayesian NMF.

The ability of Stein discrepancies to assess the quality of any collection of particles \citep{gorham2015measuring} has resulted in large recent interest in other ways to create collections of samples \citep{oates2017control, liu2016stein}.  \citet{liu2016kernelized} and \citet{chwialkowski2016kernel} showed that kernelized Stein discrepancy could be computed analytically in Reproducing Kernel Hilbert Spaces (RKHS); \citet{pu2017stein} and \citet{wanglearning} use neural networks instead.  \citet{ranganath2016operator} establish the Stein discrepancy as a valid variational objective.  To our knowledge, Stein discrepancy-based posterior approximation has not been applied to NMF, and yet, we see that it allows us to leverage existing non-Bayesian approaches to characterize these multi-modal posteriors. In our work, the Dirichlet prior on the columns of the basis matrix $A$ is important to ensure that we avoid a known saddle point of the zero factorization (from likelihood term) that yields a corresponding zero for the score function.

Our $Q$-Transform approach to finding multiple optima is most similar to \citet{rovckova2016fast} and \citet{paatero1994positive},

who use rotations to find solutions to a single matrix factorization problem that are sparse and non-negative respectively.  In contrast, we use rotations to find multiple non-negative solutions, and also demonstrate how these rotations can be \emph{re-used} for transfer learning.

\section{Conclusion}
In this work, we presented a novel transfer learning-based approach to
posterior estimation in Bayesian NMF.  Simply creating collections of
factorizations via random restarts or our $Q$-Transform initializations,
and then weighing them, produces diverse collections that
approximate the posterior well (the NNDSVDar-based methods fail to
produce diverse collections for posterior estimation).  In contrast, the
functional gradient descent of SVGD and direct gradients of Stein
discrepancy (DSGD) perform worse to the collection-based approaches,
 requiring more time and also limiting the user to specify in advance the number of
factorizations. Hamiltonian Monte Carlo also suffers from difficulties in exploring the posterior space, something random initializations are well suited to. Our transfer learning approach consistently produces the highest quality posterior approximations. 

Through $Q$-Transform, we introduce a way to speed-up the process
of finding multiple diverse NMFs.  The discovery that $Q$-Transform
matrices can transfer from synthetic to multiple real datasets is
exciting and also suggests interesting questions for further
research. For example, what is the theoretical nature of the similarities between principal eigenspaces
of different non-negative matrices and the relation between their SVD
and NMF bases? And, how does the synthetic data generation process used to obtain $Q$-Transform matrices impact the initializations and the effectiveness of the $Q$-Transform algorithm in general? 

More broadly, our qualitative results demonstrate that even relatively
simple models, such as NMF, can have multiple optima that are
comparable under the objective function but have large variation in
how well they explain different portions of the data---or how they
perform on different downstream tasks.  Thus, it is important to be
able to compute these posteriors efficiently.

\acks{We would like to thank Andreas Krause, Mike Hughes, Melanie Pradier, Nick Foti, Omer Gottesman and other members of the Dtak lab at Harvard for many helpful conversations and feedback. MAM and FDV acknowledge support from AFOSR FA 9550-17-1-0155.}

\newpage

\vskip 0.2in
\bibliography{Transfer_NMF.bib}

\clearpage

\appendix
\section*{Appendix A: Quality of factorizations}

\begin{figure}[b]
\centering
\includegraphics[width=.3\textwidth]{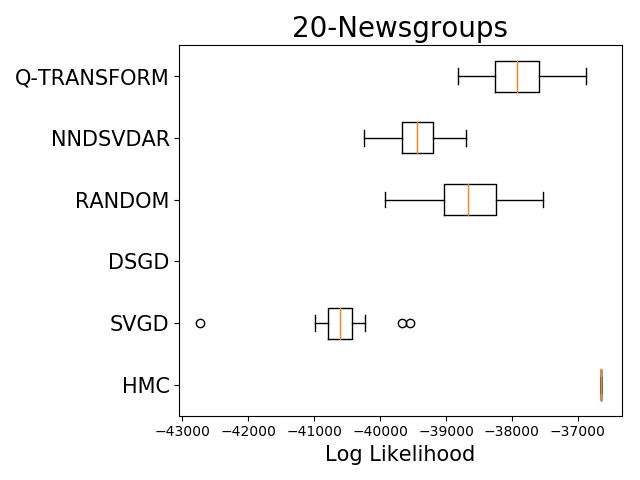}\quad
\includegraphics[width=.3\textwidth]{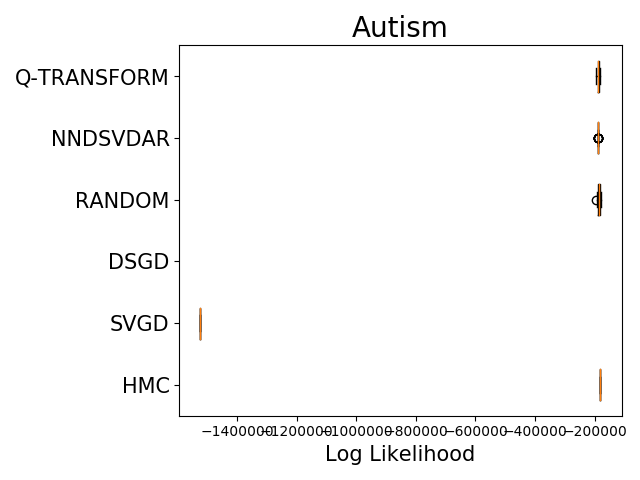}\quad
\includegraphics[width=.3\textwidth]{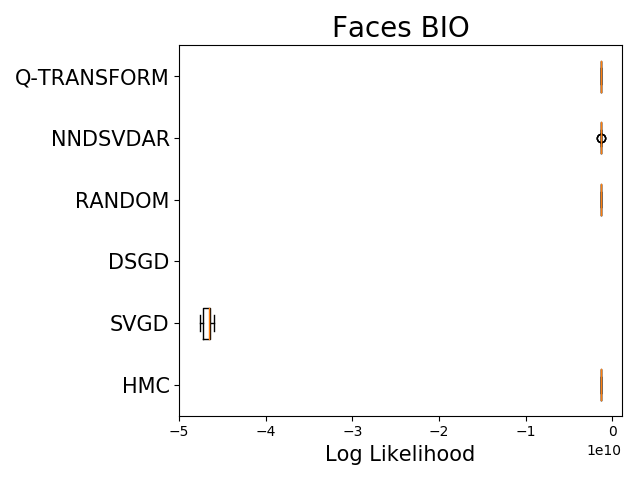}

\medskip

\includegraphics[width=.3\textwidth]{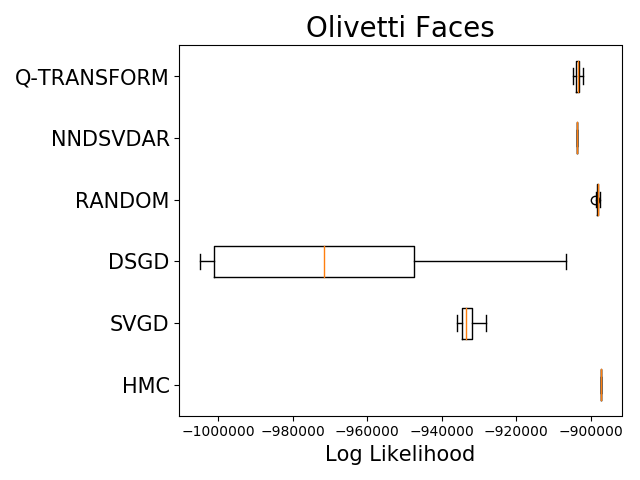}\quad
\includegraphics[width=.3\textwidth]{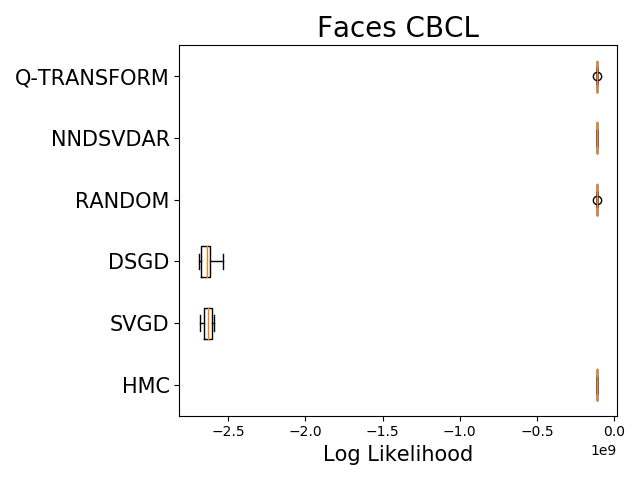}\quad
\includegraphics[width=.3\textwidth]{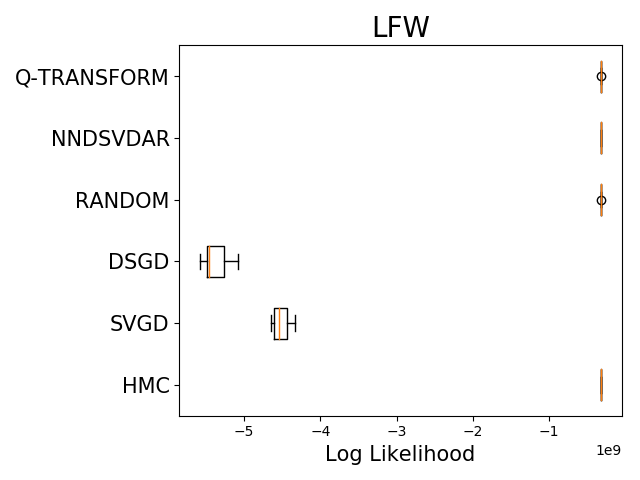}

\medskip 

\includegraphics[width=.3\textwidth]{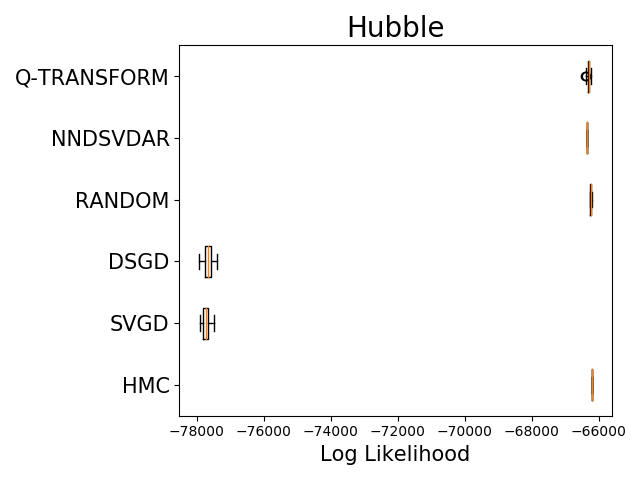}\quad
\includegraphics[width=.3\textwidth]{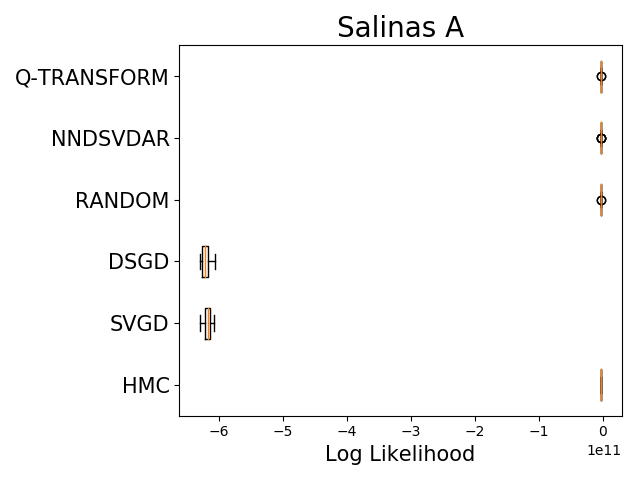}\quad
\includegraphics[width=.3\textwidth]{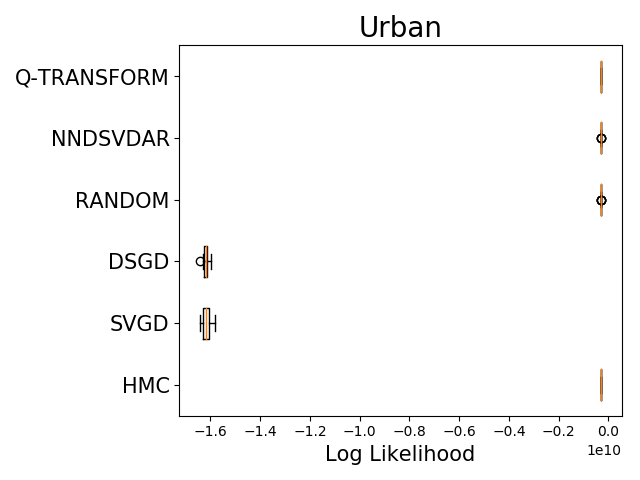}

\medskip 
\medskip 

\includegraphics[width=.3\textwidth]{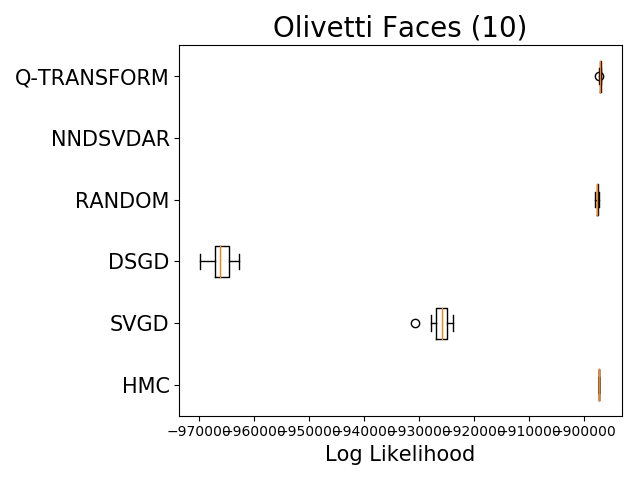}\quad
\includegraphics[width=.3\textwidth]{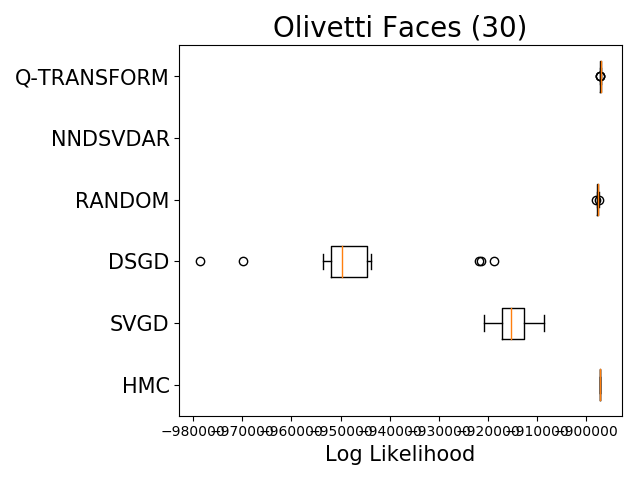}\quad
\includegraphics[width=.3\textwidth]{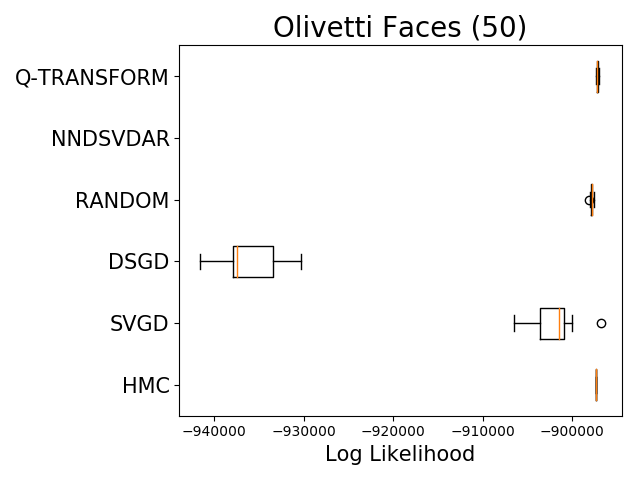}

\caption{The joint likelihood of factorizations shows that SVGD and DSGD generally produce the worst quality factorizations. HMC, NNDSVDar, Random and $Q$-Transform produce high quality factorizations.}
\label{fig:llik}
\end{figure}

\begin{figure}[b]
\centering
\includegraphics[width=.3\textwidth]{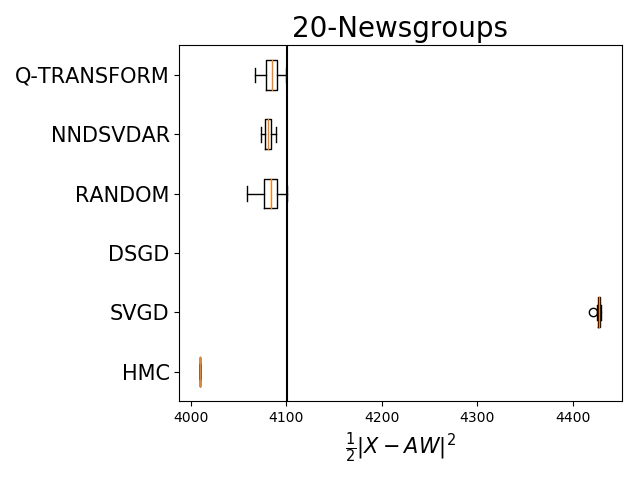}\quad
\includegraphics[width=.3\textwidth]{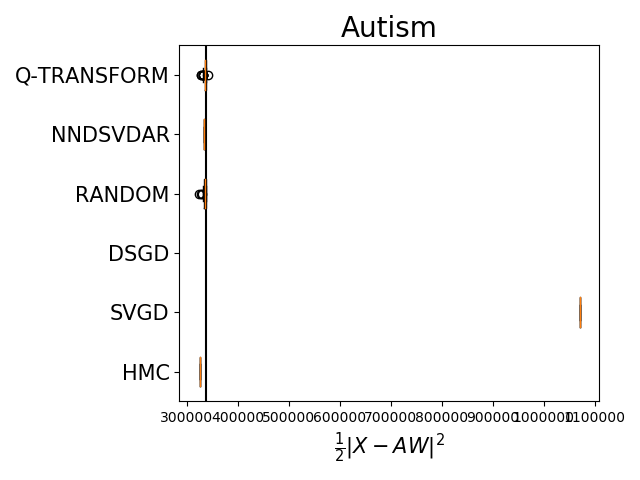}\quad
\includegraphics[width=.3\textwidth]{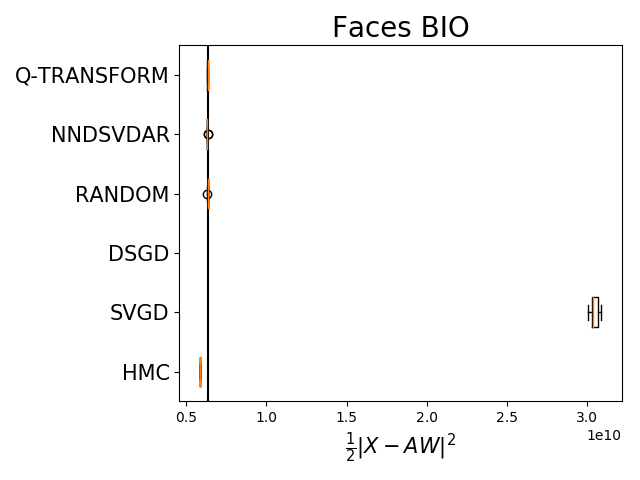}

\medskip

\includegraphics[width=.3\textwidth]{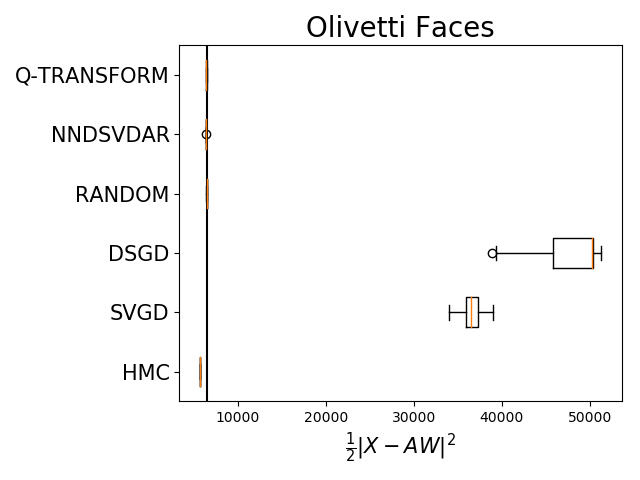}\quad
\includegraphics[width=.3\textwidth]{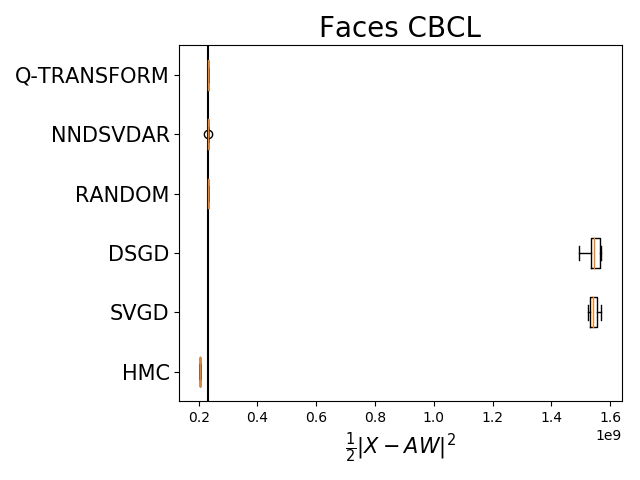}\quad
\includegraphics[width=.3\textwidth]{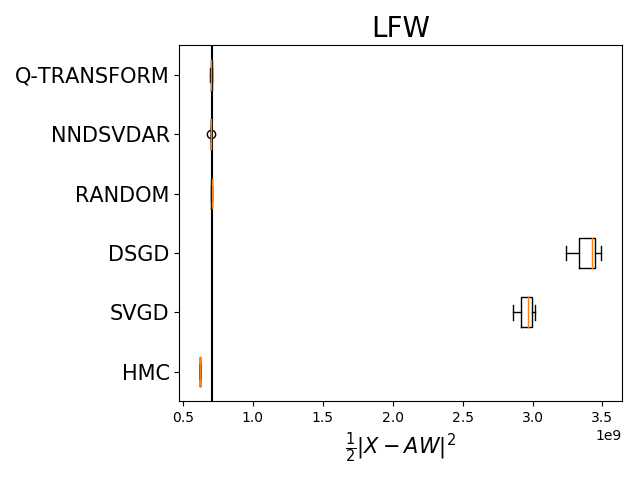}

\medskip 

\includegraphics[width=.3\textwidth]{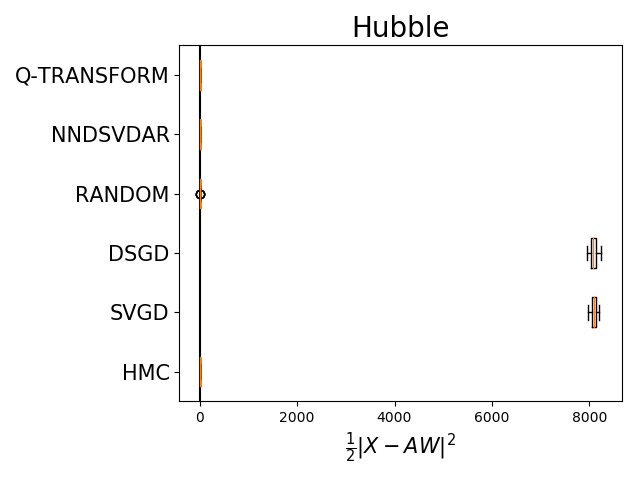}\quad
\includegraphics[width=.3\textwidth]{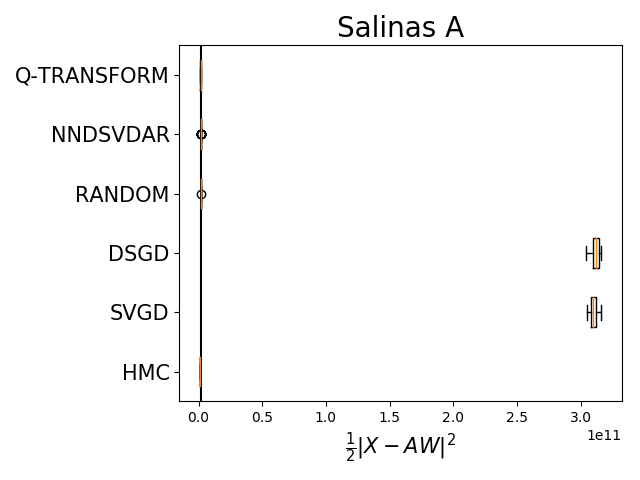}\quad
\includegraphics[width=.3\textwidth]{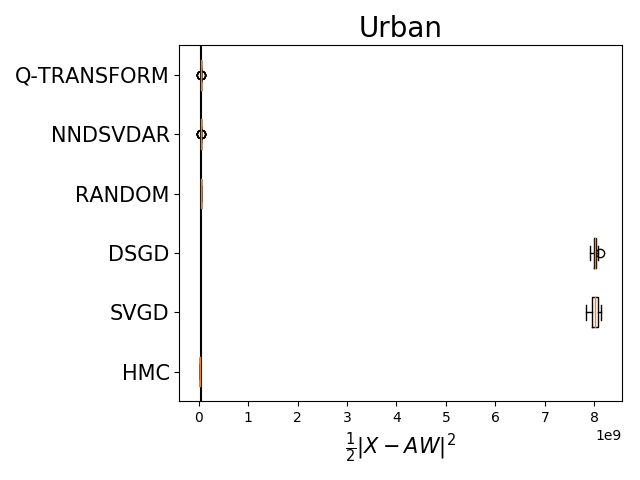}

\medskip 

\includegraphics[width=.3\textwidth]{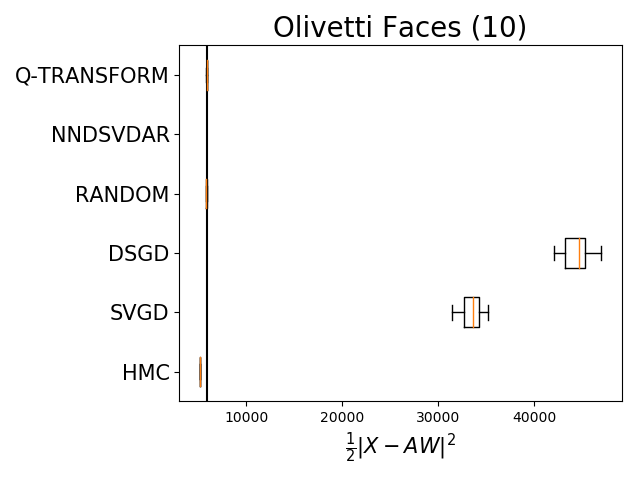}\quad
\includegraphics[width=.3\textwidth]{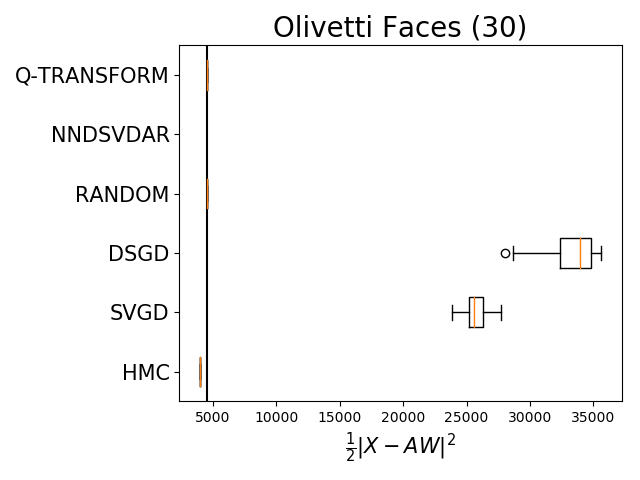}\quad
\includegraphics[width=.3\textwidth]{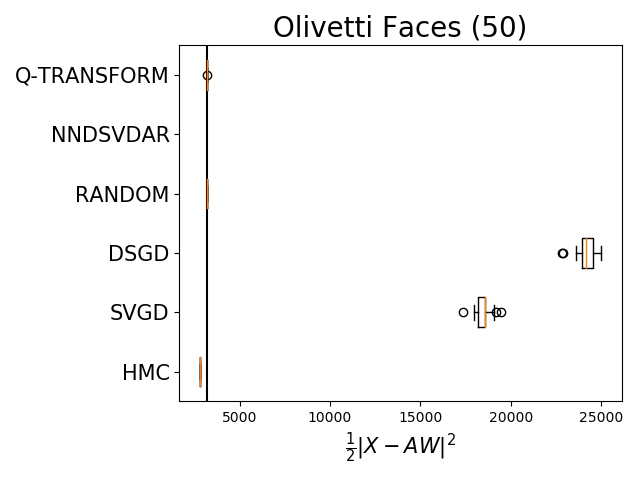}

\caption{The reconstruction error of the factorizations shows that SVGD and DSGD are typically unable to find factorization parameters that meet the threshold quality (black line) for useful factorizations. The other approaches consistently produce factorizations that meet this minimum quality requirement.}
\label{fig:fro}
\end{figure}

We measure the quality of factorizations in terms of the log of the joint likelihood (figure~\ref{fig:llik}) as well the Frobenius NMF objective (figure~\ref{fig:fro}). We see that both quality measures are in agreement with each other. $Q$-Transform, Random, NNDSVDar and HMC produce high quality factorizations. The initialization-based approaches all work well as they use specialized NMF algorithms that are designed to find high quality factorizations. HMC was given a warm start with a high likelihood initialization and the chain continues to stay in high likelihood regions. The remaining gradient-based approaches for optimizing a collection of particles (SVGD and DSGD) fail to produce high quality factorizations. This is indicative of the need for specialized NMF algorithms designed to work with the constraints and structure of the NMF problem and highlight how difficult it is to apply a naive gradient descent approach for finding NMFs. The objective function in DSGD is the Stein discrepancy, but $Q$-Transform (and almost all other baselines) perform better in that objective than DSGD.  

\section*{Appendix B: Diversity of factorizations}

\begin{figure}[b]
\centering
\includegraphics[width=.3\textwidth]{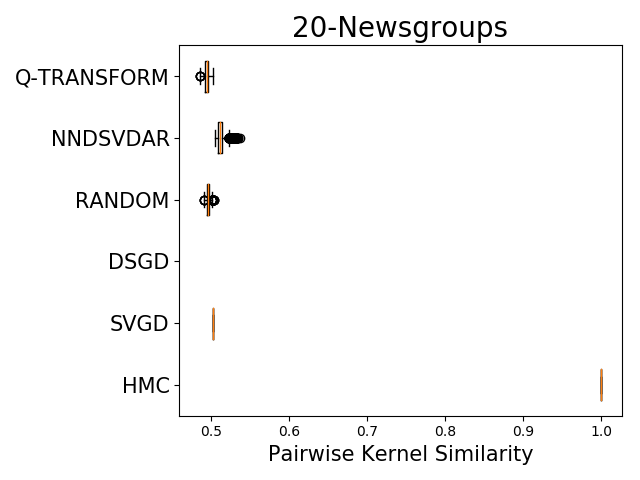}\quad
\includegraphics[width=.3\textwidth]{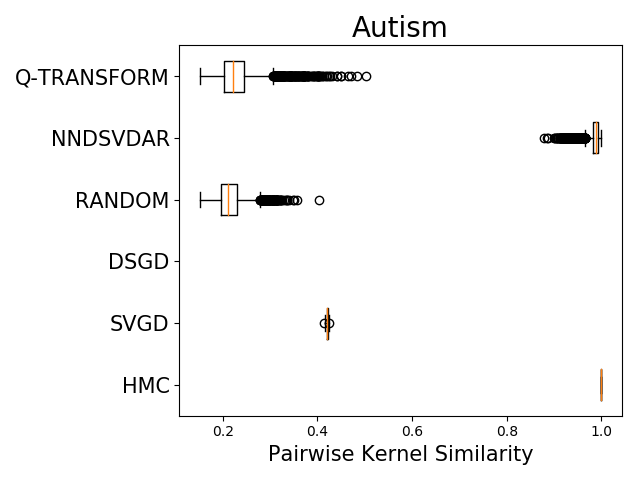}\quad
\includegraphics[width=.3\textwidth]{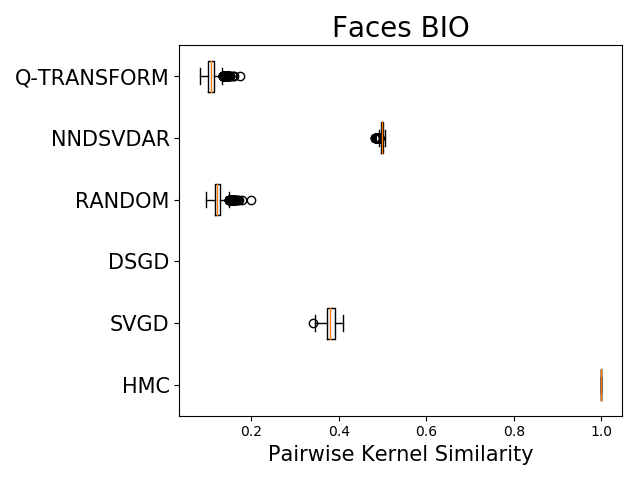}

\medskip

\includegraphics[width=.3\textwidth]{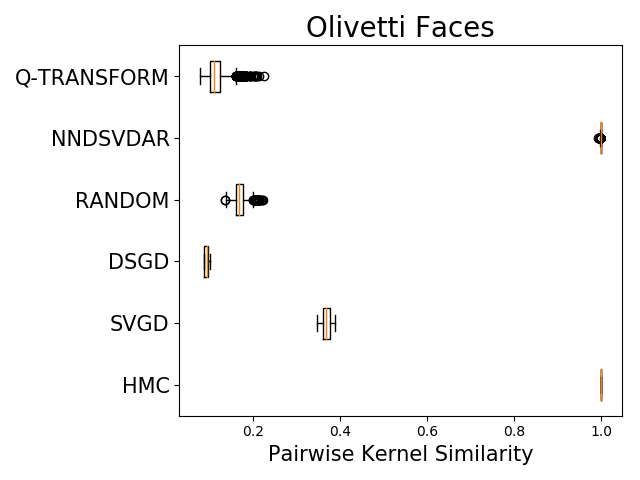}\quad
\includegraphics[width=.3\textwidth]{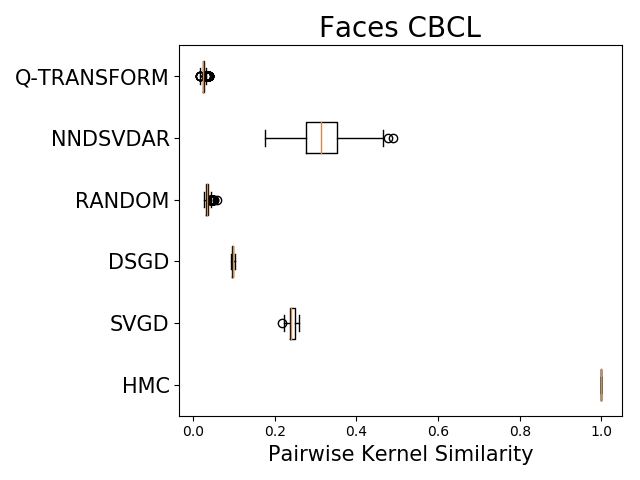}\quad
\includegraphics[width=.3\textwidth]{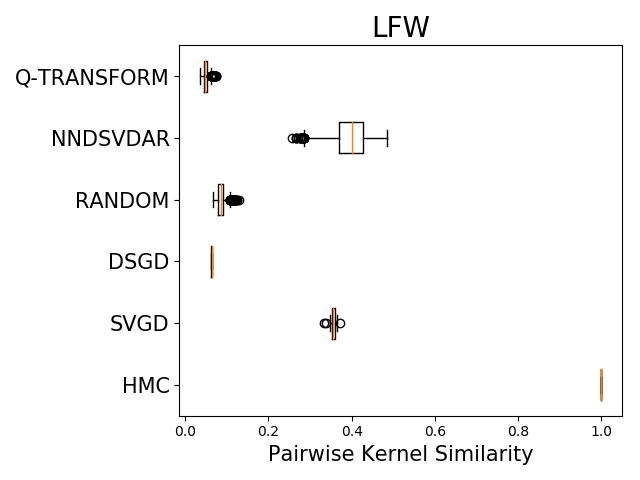}

\medskip 

\includegraphics[width=.3\textwidth]{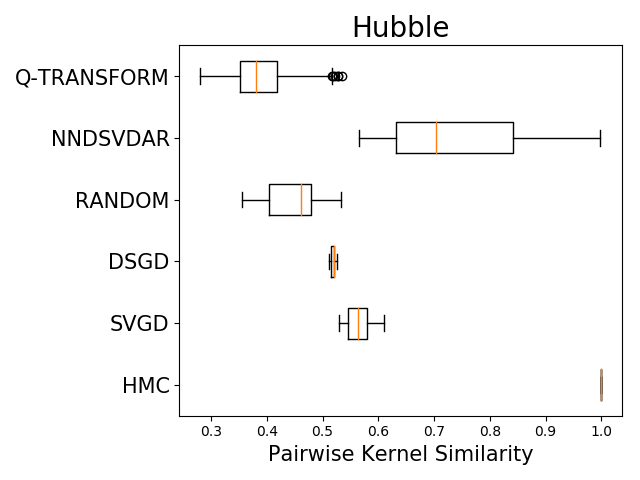}\quad
\includegraphics[width=.3\textwidth]{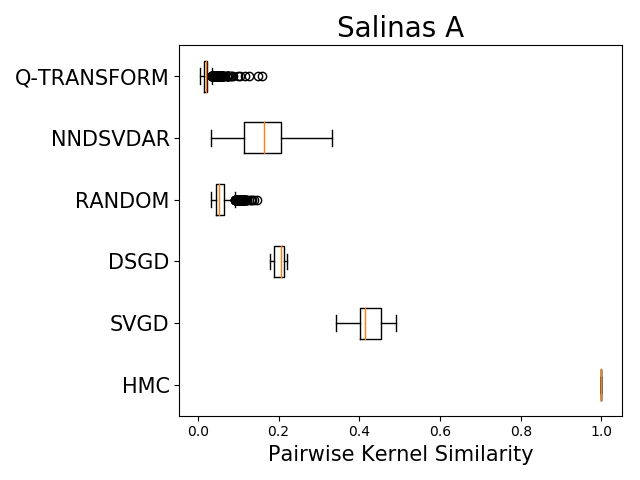}\quad
\includegraphics[width=.3\textwidth]{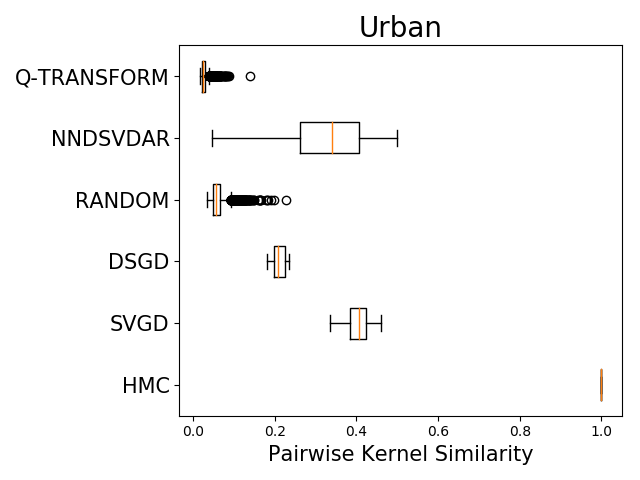}

\medskip 

\includegraphics[width=.3\textwidth]{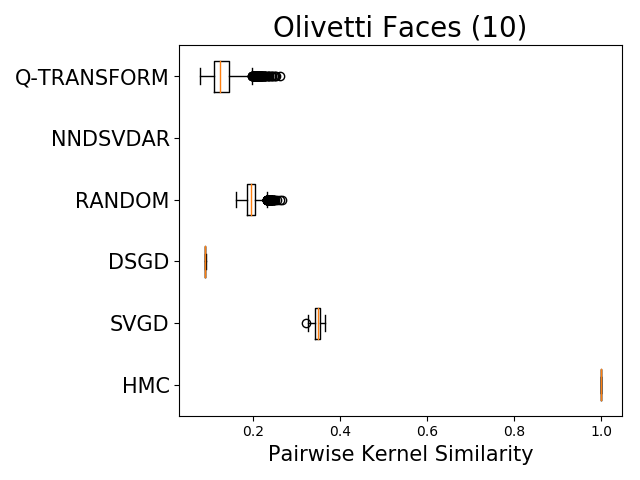}\quad
\includegraphics[width=.3\textwidth]{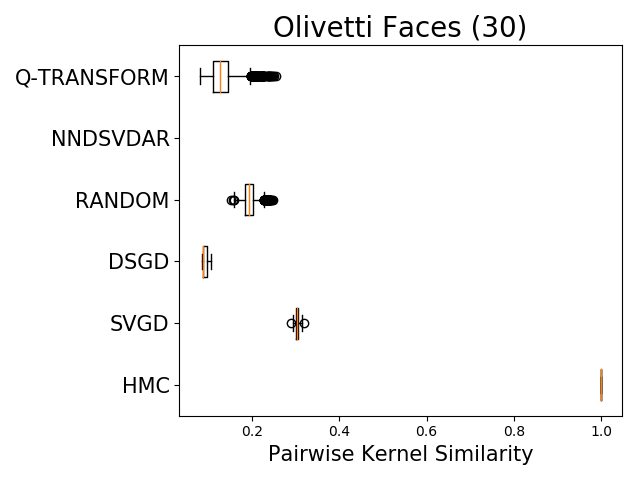}\quad
\includegraphics[width=.3\textwidth]{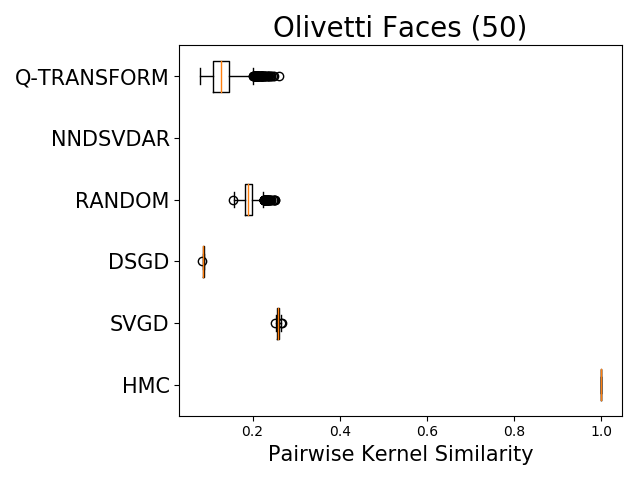}

\caption{The kernel similarity indicates that factorization collections obtained by HMC are most similar indicating that the HMC chain is only exploring a small region of the posterior. In many cases NNDSVDar factorizations are also very similar. $Q$-Transform and Random are the only algorithms that produce factorizations of high quality that are not similar.}
\label{fig:kernel}
\end{figure}

\begin{figure}[]
\centering
\includegraphics[width=.3\textwidth]{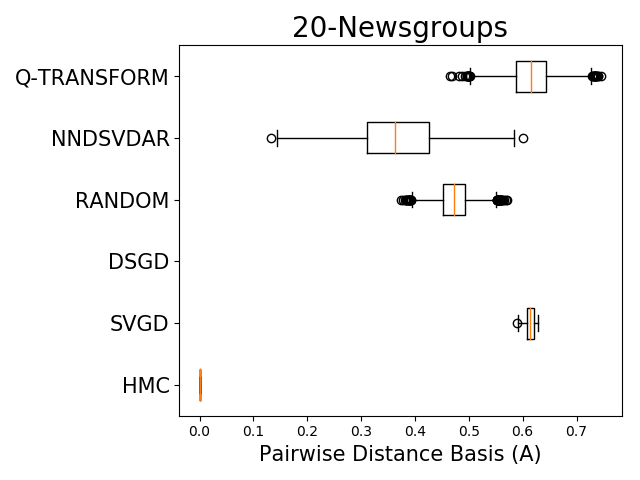}\quad
\includegraphics[width=.3\textwidth]{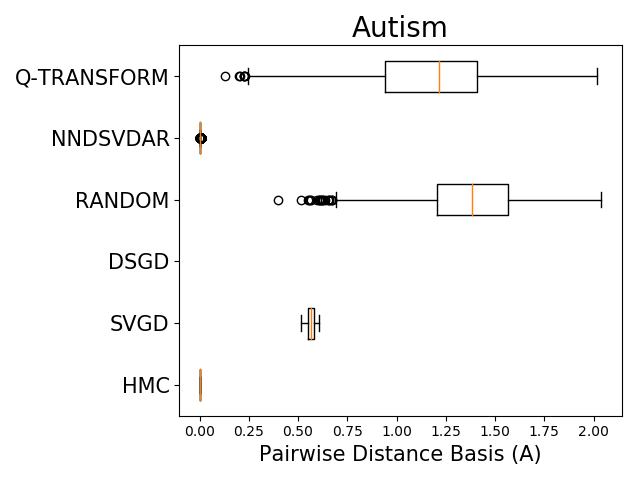}\quad
\includegraphics[width=.3\textwidth]{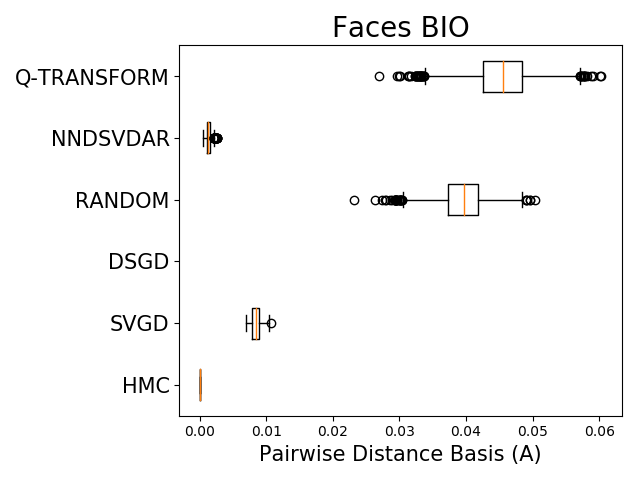}

\medskip

\includegraphics[width=.3\textwidth]{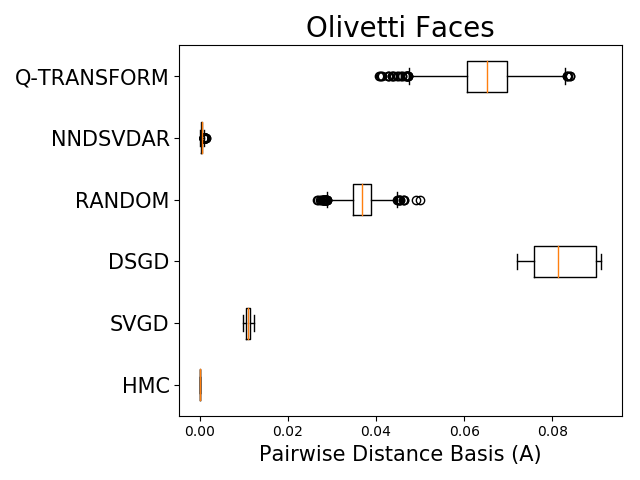}\quad
\includegraphics[width=.3\textwidth]{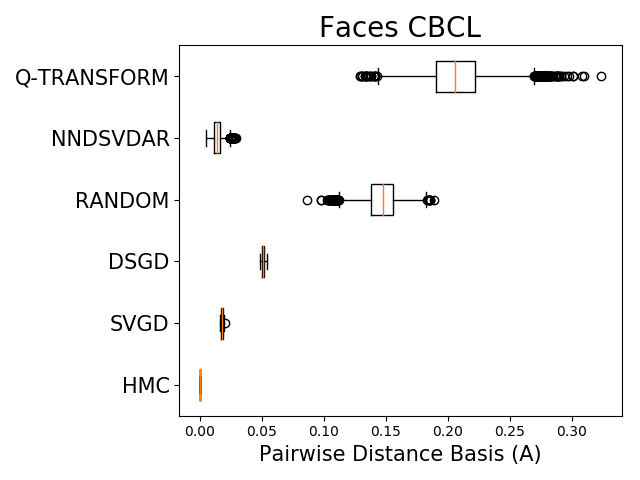}\quad
\includegraphics[width=.3\textwidth]{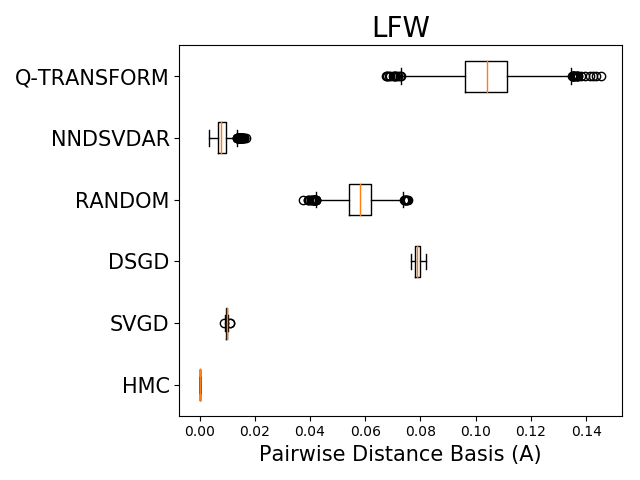}

\medskip 

\includegraphics[width=.3\textwidth]{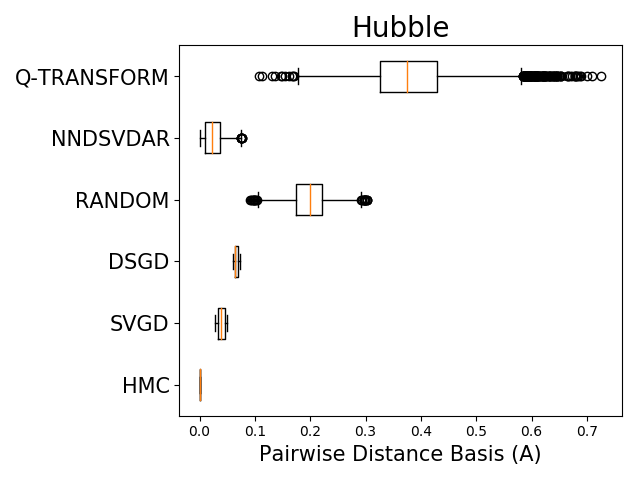}\quad
\includegraphics[width=.3\textwidth]{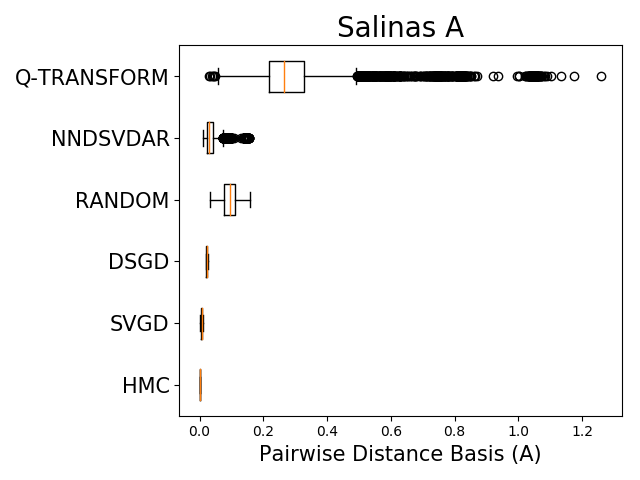}\quad
\includegraphics[width=.3\textwidth]{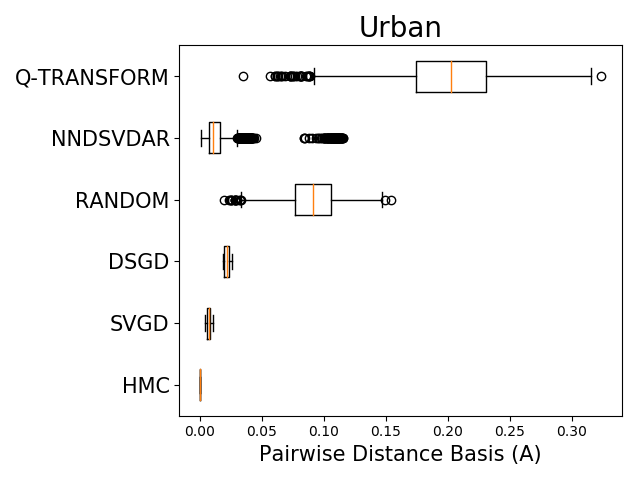}

\medskip 

\includegraphics[width=.3\textwidth]{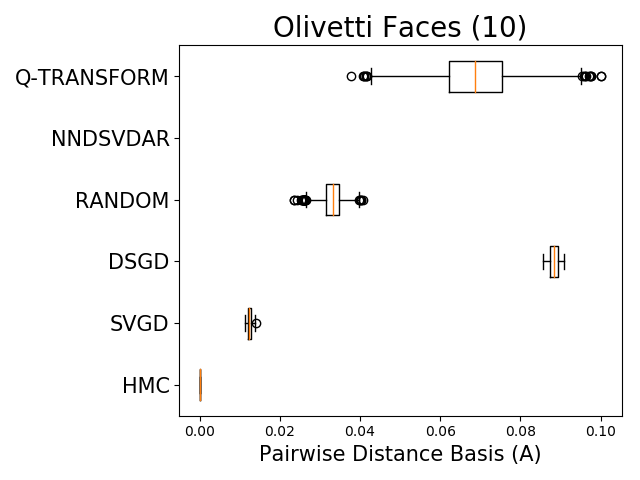}\quad
\includegraphics[width=.3\textwidth]{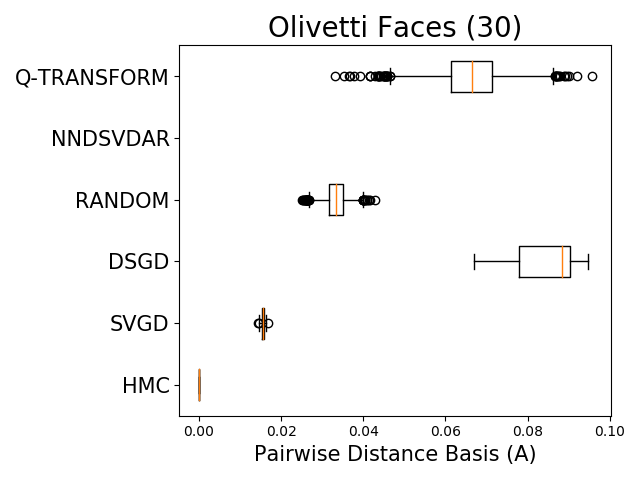}\quad
\includegraphics[width=.3\textwidth]{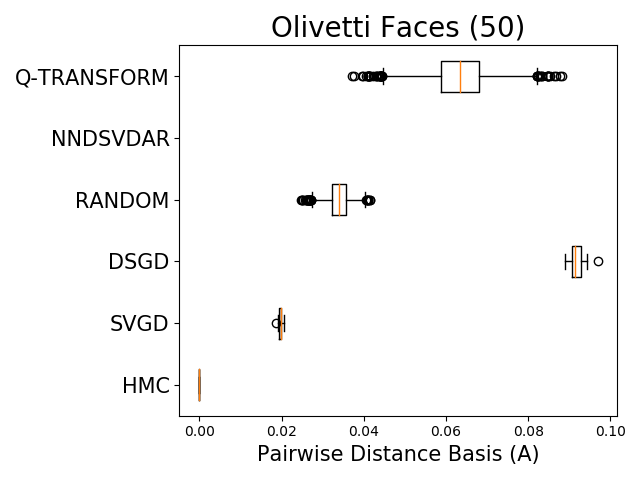}

\caption{The pairwise distance between basis matrices shows that factorization collections obtained by HMC are most similar indicating that the HMC chain is only exploring a small region of the posterior. In many cases NNDSVDar factorizations are also very similar. $Q$-Transform and Random are the only algorithms that produce factorizations of high quality that are different.}
\label{fig:basis}
\end{figure}

\begin{figure}[]
\centering
\includegraphics[width=.3\textwidth]{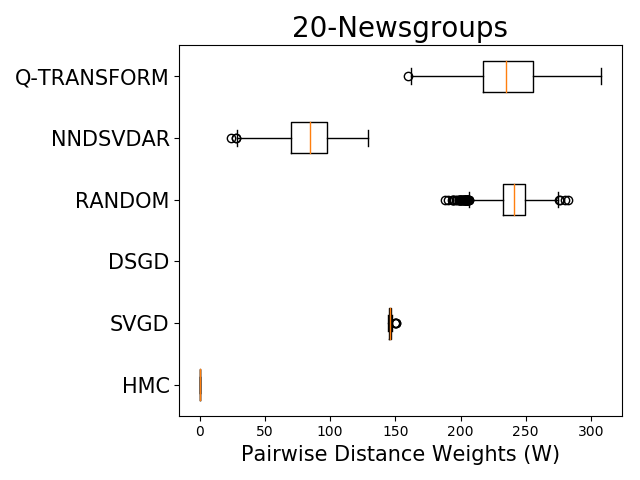}\quad
\includegraphics[width=.3\textwidth]{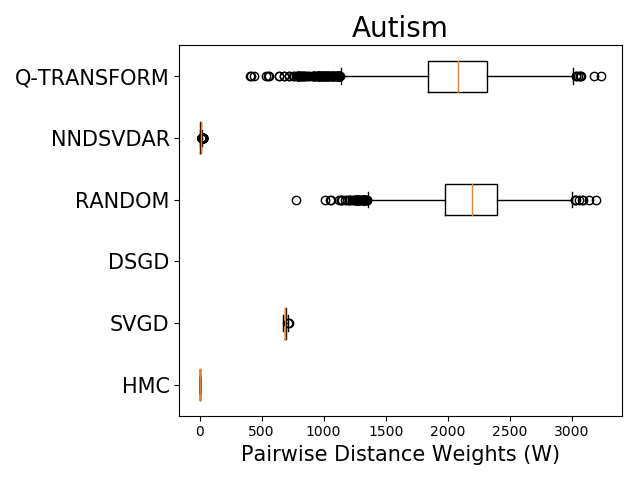}\quad
\includegraphics[width=.3\textwidth]{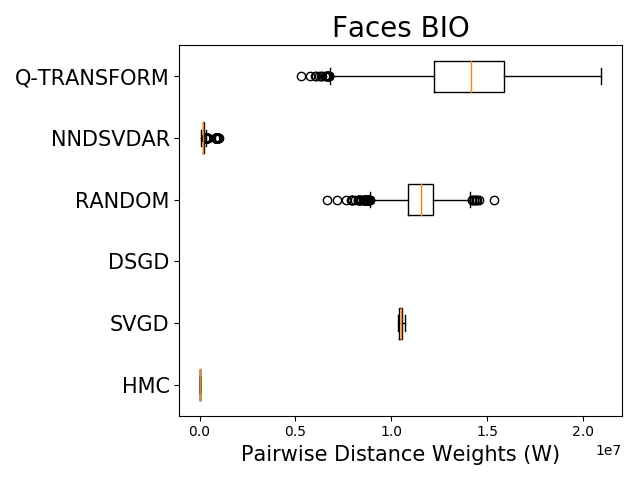}

\medskip

\includegraphics[width=.3\textwidth]{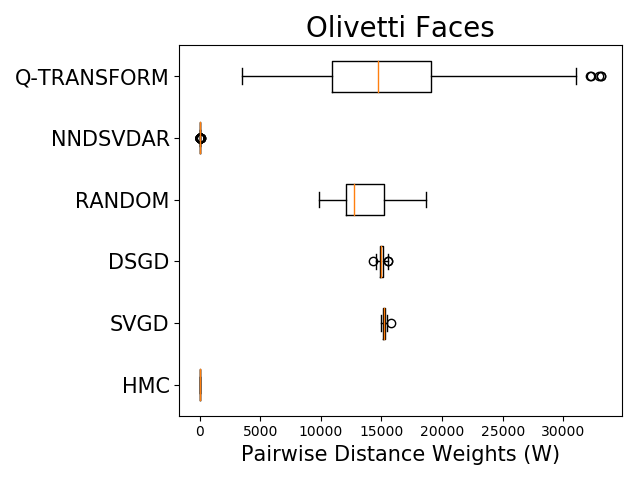}\quad
\includegraphics[width=.3\textwidth]{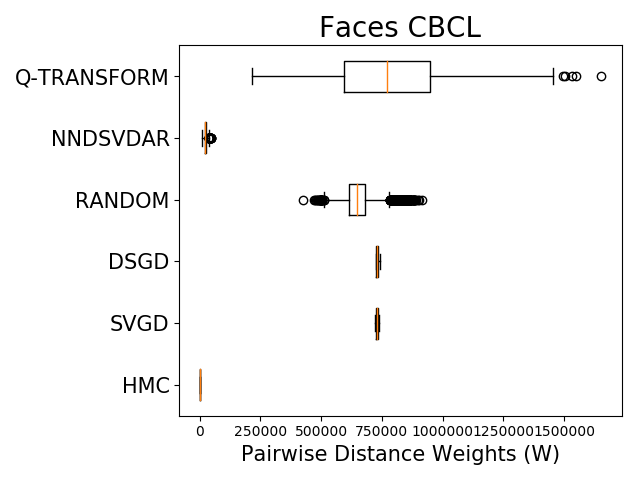}\quad
\includegraphics[width=.3\textwidth]{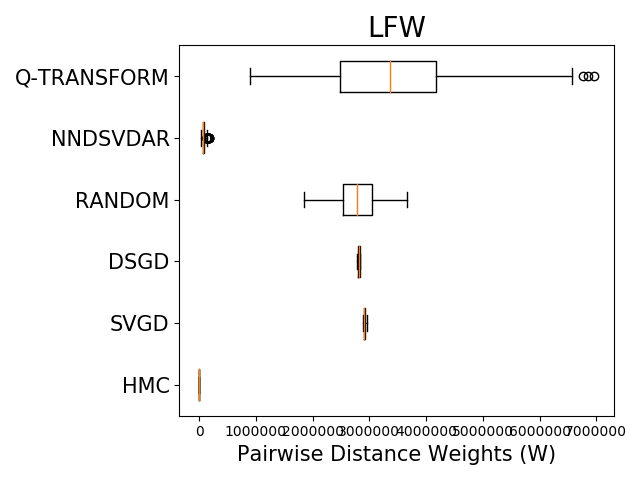}

\medskip 

\includegraphics[width=.3\textwidth]{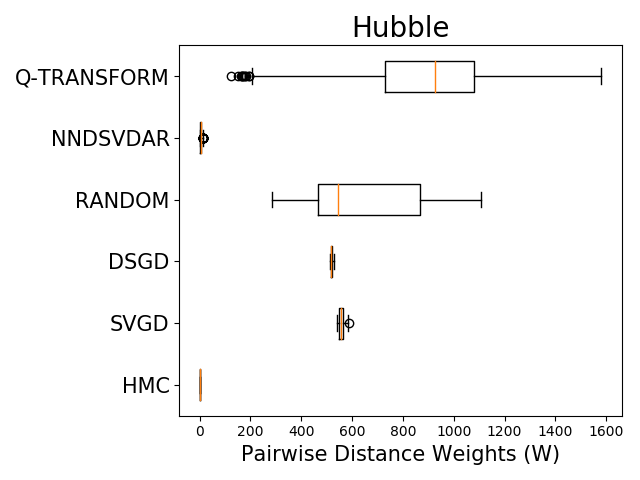}\quad
\includegraphics[width=.3\textwidth]{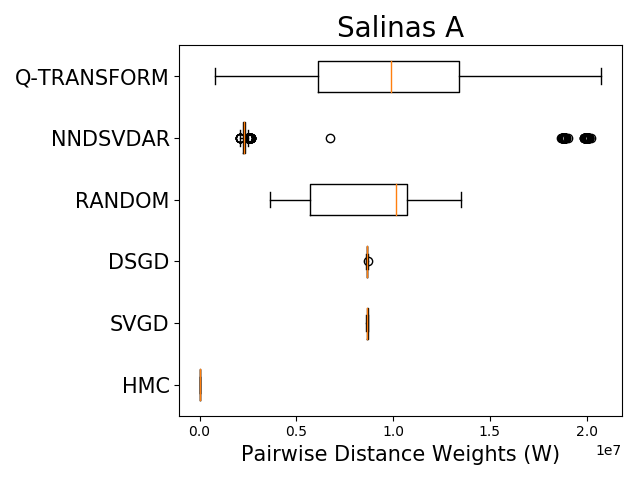}\quad
\includegraphics[width=.3\textwidth]{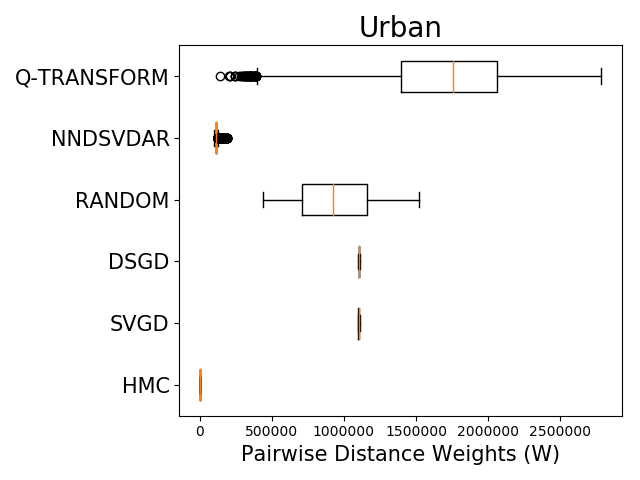}

\medskip 

\includegraphics[width=.3\textwidth]{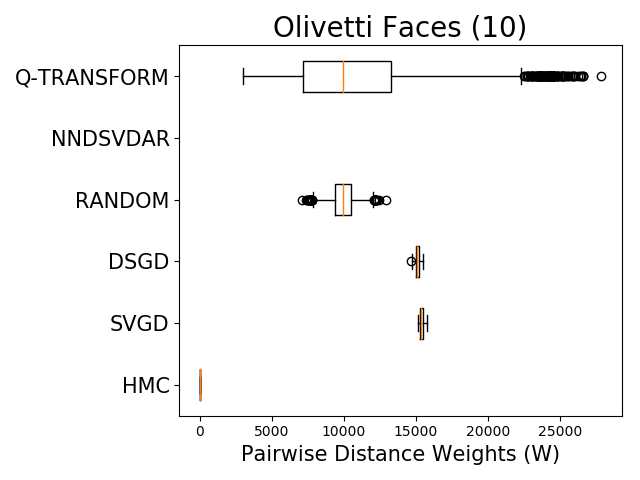}\quad
\includegraphics[width=.3\textwidth]{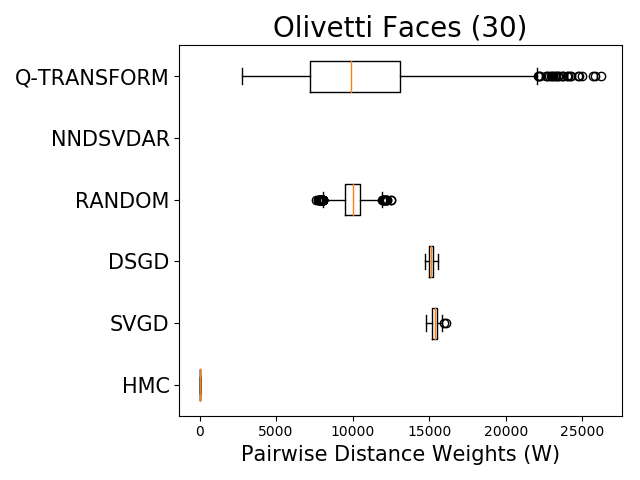}\quad
\includegraphics[width=.3\textwidth]{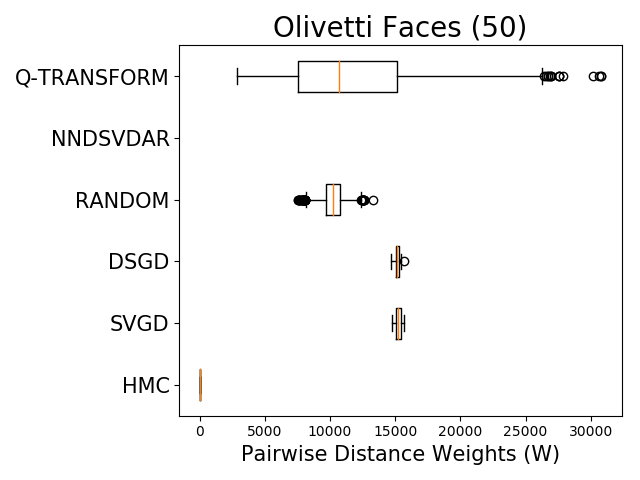}
\caption{The pairwise distance between weights matrices shows that factorization collections obtained by HMC are most similar indicating that the HMC chain is only exploring a small region of the posterior. In many cases NNDSVDar factorizations are also very similar. $Q$-Transform and Random are the only algorithms that produce factorizations of high quality that are different.}
\label{fig:weights}
\end{figure}

We take a closer look at the diversity of factorizations produced by our $Q$-Transform and competing baseline approaches. Similarity of factorizations is measured by the kernel (equation~\ref{eqn:my_imq}) for the base RKHS (figure~\ref{fig:kernel}) used in evaluating the Stein discrepancy and by pairwise distances between basis matrices (figure~\ref{fig:basis}) and weights matrices (figure~\ref{fig:weights}). The different diversity metrics tell the same story. The HMC chain exhibits the least exploration of the factorization space followed by NNDSVDar. SVGD, DSGD, Random and $Q$-Transform all exhibit the higher amounts of exploration in the factorization space, but the exploration in SVGD and DSGD does not correspond to high likelihood regions of the posterior therefore it is unimportant.
 
\end{document}